\documentclass[sigconf]{acmart}
\usepackage{caption}        % 表格标题格式优化
\usepackage{multirow}       % 多行列支持
\usepackage{pifont}         % 提供ding命令支持
\usepackage{xcolor}
\definecolor{darkgreen}{RGB}{0, 100, 0} % 通过RGB定义深绿色
\AtBeginDocument{%
  }
\usepackage{pifont} % for \ding symbols
\usepackage{multirow}
\usepackage[table]{xcolor}  % 加上 table 选项支持表格着色
\definecolor{mred}{RGB}{238, 34, 12}
\definecolor{mgreen}{RGB}{1, 127, 0}
\definecolor{mblue}{RGB}{0, 77, 158}
\newcommand{\mredbf}[1]{\textcolor{mred}{\textbf{#1}}}
\newcommand{\mbluebf}[1]{\textcolor{mblue}{\textbf{#1}}}
\usepackage{amsmath}
\usepackage{arydshln} % 支持虚线
\usepackage{makecell}
\usepackage{enumitem}
\setcopyright{acmlicensed}
\copyrightyear{2026}
\acmYear{2026}
\acmDOI{XXXXXXX.XXXXXXX}

\acmConference[MM '26]{ACM Multimedia}{November 10--14 2026}{Rio de Janeiro, Brazil}
\acmISBN{978-1-4503-XXXX-X/26/11}

\begin{document}

% ===== Title (draft; inspired by LMM4Edit + VELA-style evaluation framing) =====
\title{ITIScore: An Image-to-Text-to-Image Rating Framework for the Image Captioning Ability of MLLMs}

% ===== Authors (replace with your own; keep structure) =====
% For anonymous review, you can comment out authors entirely when using anonymous option.
\author{Zitong Xu\textsuperscript{1}, Huiyu Duan\textsuperscript{1}\textsuperscript{†}, Shengyao Qin\textsuperscript{2}, Guangyu Yao\textsuperscript{2}, Guangji Ma\textsuperscript{2}, \\Xiongkuo Min\textsuperscript{1}\textsuperscript{†}, Ke Gu\textsuperscript{3}, Guangtao Zhai\textsuperscript{1}\textsuperscript{†}, Patrick Le Callet\textsuperscript{4}}
\author{\textsuperscript{1}Shanghai Jiao Tong University,
\textsuperscript{2}University of Electronic and Science Technology of China, \\
\textsuperscript{3}Beijing University of Technology, \textsuperscript{4}Institut Universitaire de France (IUF), University of Nantes \\† Corresponding authors.\\
$\{$xuzitong, huiyuduan, minxiongkuo, zhaiguangtao$\}$@sjtu.edu.cn}
% \affiliation{%
%   \institution{Shanghai Jiao Tong University}
%   \city{Shanghai}
%   \country{China}
% }
% \email{author@domain.com}
% \acmSubmissionID{385}
% If multiple authors:
% \author{Second Author}
% \affiliation{%
%   \institution{Your Institution}
%   \city{City}
%   \country{Country}
% }
% \email{second@domain.com}
\settopmatter{printacmref=false}
\renewcommand{\shortauthors}{}

% ===== Abstract =====
\begin{abstract}
Recent advances in multimodal large language models (MLLMs) have greatly improved image understanding and captioning capabilities. However, existing image captioning benchmarks typically suffer from limited diversity in caption length, the absence of recent advanced MLLMs, and insufficient human annotations, which potentially introduces bias and limits the ability to comprehensively assess the performance of modern MLLMs. To address these limitations, we present a new large-scale \underline{i}mage \underline{c}aptioning \underline{b}enchmark, termed, \textbf{ICBench}, which covers 12 content categories and consists of both short and long captions generated by 10 advanced MLLMs on 2K images, resulting in 40K captions in total. We conduct extensive human subjective studies to obtain mean opinion scores (MOSs) across fine-grained evaluation dimensions, where short captions are assessed in terms of fluency, relevance, and conciseness, while long captions are evaluated based on fluency, relevance, and completeness. Furthermore, we propose an automated evaluation metric, \textbf{ITIScore}, based on an image-to-text-to-image framework, which measures caption quality through reconstruction consistency. Experimental results demonstrate strong alignment between our automatic metric and human judgments, as well as robust zero-shot generalization ability on other public captioning datasets. Both the dataset and model will be released upon publication. 
\end{abstract}

% ===== CCS Concepts (replace via https://dl.acm.org/ccs) =====


% ===== Keywords =====

\begin{teaserfigure}
\centering
  \includegraphics[width=1\textwidth]{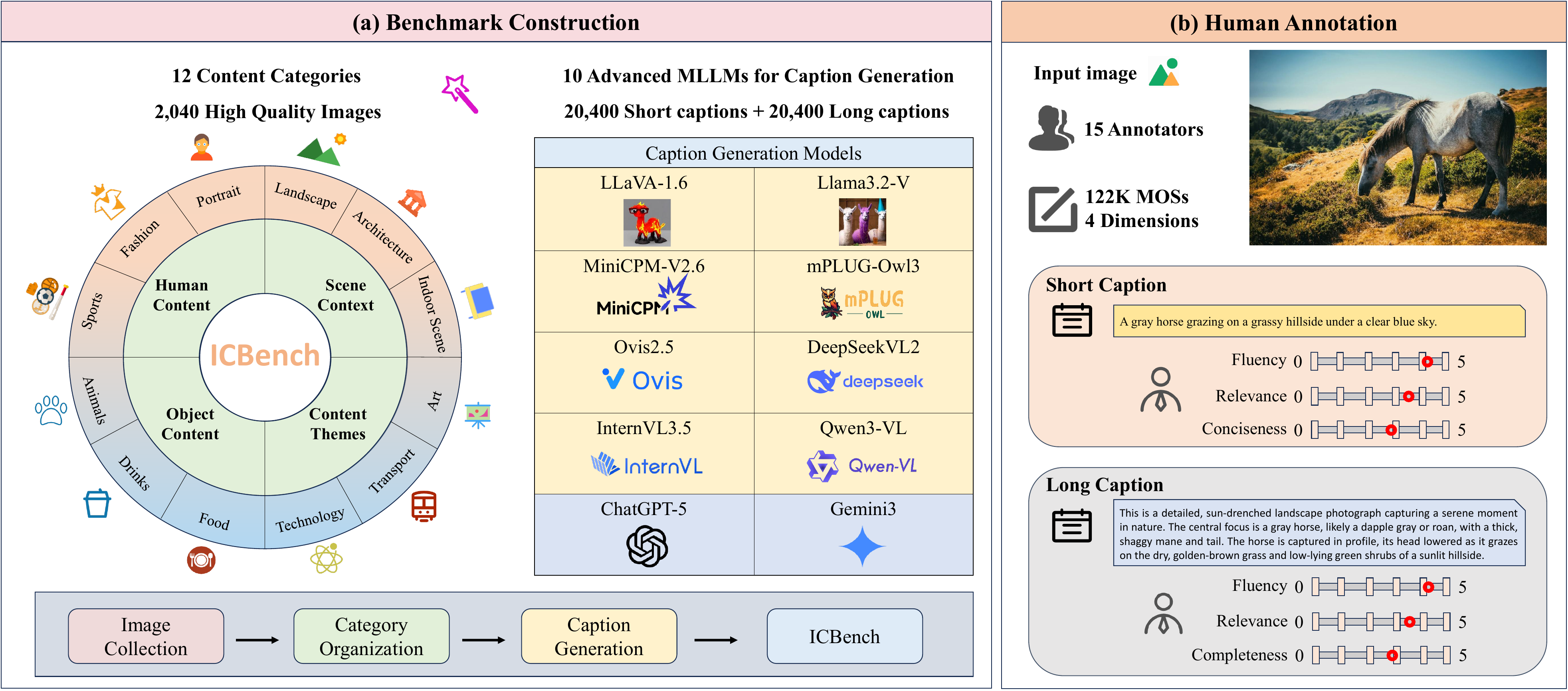}
  \caption{Overview of our ICBench. (a) We first collect 2,040 source images across 12 fine-grained tasks. Then 10 advanced MLLMs are applied to generate 20,400 short captions and 20,400 long captions. (b) 122K MOSs are collected from 15 annotators across 4 evaluation dimensions.}
  \label{overall}
\end{teaserfigure}

\maketitle
% \setlength{\textfloatsep}{5pt plus 0.5pt minus 0.5pt}
% \setlength{\dbltextfloatsep}{5pt plus 0.5pt minus 0.5pt}
% \setlength{\dblfloatsep}{5pt plus 0.5pt minus 0.5pt}
% \setlength{\intextsep}{5pt plus 0.5pt minus 0.5pt}
% \setlength{\abovecaptionskip}{5pt plus 0.5pt minus 0.5pt}
% \setlength{\belowcaptionskip}{5pt plus 0.5pt minus 0.5pt}
% \setlength{\abovedisplayskip}{2pt}
% \setlength{\belowdisplayskip}{2pt}
% % \setlength{\topsep}{1.5pt}
% \titlespacing*{\section}{0pt}{7pt plus 2pt minus 2pt}{7pt plus 2pt minus 2pt} % {左边距}{上间距}{下间距}
% \titlespacing*{\subsection}{0pt}{7pt plus 2pt minus 2pt}{7pt plus 2pt minus 2pt} % {左边距}{上间距}{下间距}
% % 设置 paragraph 的格式
% \titlespacing*{\paragraph}
% {0pt}    % 左缩进
% {0.5em}    % 标题前的垂直间距
% {0.5em}    % 标题后正文与标题的间距
% \setlength{\abovedisplayskip}{4pt}   % 上方间距
% \setlength{\belowdisplayskip}{4pt}   % 下方间距
\section{Introduction}
Image captioning aims to generate natural language descriptions that faithfully reflect the content of an image \cite{Flickr8K-CF}. As a fundamental task bridging computer vision and natural language processing, it plays a crucial role in applications such as visual understanding, human-computer interaction, and multimodal reasoning \cite{wang2020overview,ohi2024multi}. With the rapid development of multimodal large language models (MLLMs), they have demonstrated strong multimodal understanding capabilities \cite{lmm4edit,xu2026edithf1mmillionscalerichhuman,xu2026manipshieldunifiedframeworkimage,harmonyiqa,liu2025moa,yang2025odi}, leading to remarkable improvements in generating fluent and detailed captions. These models are capable of producing both concise summaries and long-form captions that capture rich visual semantics \cite{vela}. However, the progress of captioning models has also exposed growing challenges of developing evaluation metrics that align well with human preference.

Moreover, automatic caption evaluation remains a longstanding challenge. Traditional metrics such as BLEU \cite{papineni2002bleu} , METEOR \cite{banerjee2005meteor}, ROUGE \cite{lin2004rouge}, and CIDEr \cite{cider} rely on lexical overlap with reference captions, which often fails to reflect semantic correctness or visual grounding. Although consensus-based metrics and vision-language metrics such as CLIP-S \cite{clipscore} attempt to address this limitation, they still exhibit limited correlation with human judgment, particularly for long and information-rich captions \cite{rotstein2024fusecap}. Recent approaches have explored the use of MLLMs for caption evaluation, providing more interpretable assessments across multiple perspectives \cite{chan2023clair,ohi2024multi,maeda2024vision}. Nevertheless, a comprehensive study exploring the effectiveness of MLLMs in evaluating image captioning tasks is still lacking. Key limitations lies in the evaluation benchmarks includes: \textbf{(i) Caption length diversity limitation:} existing datasets typically focus on either short captions or long captions, but rarely include both, which restricts their ability to benchmark the captioning capabilities of models in terms of different lengths. \textbf{(ii) Lack of advanced MLLM:} these datasets do not include annotations for captions generated by recent state-of-the-art MLLMs, limiting their usefulness for assessing advanced models. \textbf{(iii) Few annotators:} many datasets provide annotations from only 3 to 5 annotators per image, making the results prone to bias and insufficient for developing fine-grained metrics aligned with human perception. \textbf{(iv) Potential data leakage:} the images are usually drawn from public sources, some of which may have been seen by MLLMs during pretraining, raising concerns about training-test contamination.

To address these limitations, we introduce \textbf{ICBench}, a new large-scale benchmark for systematic evaluation of image captioning capability, as shown in Figure~\ref{overall}. Unlike previous datasets that focus on a single caption style, ICBench evaluates both short captions and long captions within a unified framework. The benchmark covers 12 diverse image content categories and includes captions generated by 10 advanced MLLMs on 2,040 images, resulting in a total of 40,800 captions. We conduct extensive human subjective studies to obtain reliable MOS annotations across fine-grained evaluation dimensions. Specifically, short captions are evaluated in terms of \textit{fluency}, \textit{relevance}, and \textit{conciseness}, while long captions are assessed based on \textit{fluency}, \textit{relevance}, and \textit{completeness}. In total, the benchmark contains approximately 1.8 million human ratings, providing a comprehensive and reliable resource for caption evaluation research.

Based on this benchmark, we further explore the effectiveness of these MLLMs can also serve as tools for evaluating image captioning. Moreover, we propose an automatic caption evaluation metric called \textbf{ITIScore}, which adopts an image-to-text-to-image framework. Instead of relying solely on textual similarity, ITIScore evaluates captioning quality by measuring the reconstruction consistency between the original image and the image regenerated from the caption. A frozen MLLM backbone is used for multimodal feature fusion, with only a lightweight scoring head trained, enabling the metric to capture whether the caption preserves essential visual information and supporting richer semantic evaluation for both short and long captions. Extensive experiments demonstrate that ITIScore achieves strong alignment with human judgments on ICBench and generalizes well to other public captioning datasets.

The main contributions of this work include:
\begin{itemize}[left=0pt, labelsep=0.6em, labelwidth=0pt]
\item We introduce ICBench, a large-scale benchmark for image captioning evaluation covering both short and long captions, with 40K captions generated by 10 advanced MLLMs across 12 image categories.
\item We collect fine-grained MOSs through extensive human subjective studies, resulting in approximately 1.8M judgments across multiple evaluation dimensions.
\item We propose ITIScore, a novel image-to-text-to-image evaluation metric that measures caption quality via reconstruction consistency, using a frozen MLLM for multimodal feature fusion while training only a lightweight scoring head.
\item Extensive experiments on ICBench and other public datasets demonstrate that ITIScore achieves strong alignment with human judgments and exhibits great generalization ability.
\end{itemize}

% 2.1 Caption generation with VLM/LMMs
% 2.2 Caption evaluation metrics and protocols
% 2.3 Multi-model benchmarking and category-wise diagnosis
% TODO
\section{Related Work}

\subsection{Image Captioning}
Image captioning has long been studied using encoder-decoder formulations that map visual representations to natural language descriptions~\cite{xu2023deep,vinyals2015show}. Early captioning models evolve from
template-based and recurrent neural networks~\cite{vinyals2015show} to Transformer-based architectures \cite{cornia2020meshed, xu2015show,pan2020x}.
Region-level modeling and stronger visual grounding, exemplified by bottom-up and top-down attention, have substantially advanced caption quality and controllability~\cite{anderson2018bottom}.
With the rise of large-scale vision-language pretraining, unified pretraining frameworks strengthen open-domain caption generation and transfer~\cite{li2022blip,wang2022git}.
More recently, instruction-following MLLMs have made detailed, attribute-rich descriptions increasingly practical, while also exposing new failure modes where captions remain plausible but may omit key evidence or introduce subtly incorrect attributes~\cite{vela,spice,clipscore,cheng2025caparena,sarto2025image}.
% These observations motivate evaluation settings that go beyond short, generic descriptions and explicitly characterize caption quality along content-critical dimensions, especially for long, information-dense captions~\cite{vela,spice}.
% In this context, our work is positioned as a general framework for studying long captions under broader, multi-task usage scenarios, aiming for more comprehensive coverage and stronger practical generality without relying on task-specific training for evaluation~\cite{vela,clipscore}.
\begin{table*}[t]
\centering
\caption{Comparison of image captioning evaluation benchmarks.}
 \resizebox{1\textwidth}{!}{
\begin{tabular}{lcccccccccc}
\toprule
\noalign{\vspace{-1.5pt}}
Databases& Annotations  & Images&Image Source &Short Captions & Long Captions& Models & Dimensions \\
\hline
\noalign{\vspace{1.5pt}}
Composite \cite{Composite}&59,925&3,995&Public Dataset&11,985&\color{red}{\ding{55}}&2&Relevance\\
Flick8K-Expert \cite{Flickr8K-CF}&28,320&5,664&Public Dataset&5,664&\color{red}{\ding{55}}&\color{red}{\ding{55}}&Relevance\\
Flick8K-CF \cite{Flickr8K-CF}&143,490&1,000&Public Dataset&47,830&\color{red}{\ding{55}}&\color{red}{\ding{55}}&Relevance\\
Polaris \cite{polaris}&1,048,160&13,691&Public Dataset&131,020&\color{red}{\ding{55}}&10&Relevance\\
MMHE \cite{ohi2024multi}&18,000&400&Public Dataset&1,200&\color{red}{\ding{55}}&10&Correctness, Completeness, Clarity, Fluency, and Conciseness\\
THumB \cite{kasai2022transparent}&16,000&500&Public Dataset&2,000&\color{red}{\ding{55}}&4&Precision, Recall\\
LongCap-Arena \cite{vela} & 702,450 &7,805&Public Dataset&  \color{red}{\ding{55}}&78,050& 10& Descriptiveness, Relevance, Fluency\\
% IE-Bench \cite{IEBench}& \textcolor{darkgreen}{\ding{51}}   &301&1,505&
% 22,575
% &5&11&Description, Instruction&Quality, Correspondence, Preservation&\color{red}{\ding{55}} \\
\hline
\noalign{\vspace{1.5pt}}
  \rowcolor{gray!20}  % 这里添加灰色阴影
 \textbf{ICBench (Ours)} & \textbf{1,836,000} &  \textbf{2,040}&\textbf{Web Images}&\textbf{20,400}& \textbf{
 20,400}& \textbf{10}& \textbf{Fluency, Relevance, Conciseness, Completeness}\\
 \noalign{\vspace{-1.5pt}}
\bottomrule
\label{comparison}
\end{tabular}}
\end{table*}
\subsection{Image Captioning Evaluation Dataset}
Table~\ref{comparison} provides an overview of existing image caption evaluation datasets. Traditional datasets for evaluating image captioning metrics include Composite \cite{Composite}, Flickr8K-Expert \cite{Flickr8K-CF}, Flickr8K-CF \cite{Flickr8K-CF} and Polaris \cite{polaris}. However, these datasets provide human judgments from only a single evaluation perspective. Recent studies have introduced datasets with multi-dimensional human annotations \cite{ohi2024multi,kasai2022transparent,vela}. For example, MMHE \cite{ohi2024multi} provides 4,500 human judgments for 100 images across five evaluation perspectives: \textit{Correctness}, \textit{Completeness}, \textit{Clarity}, \textit{Fluency}, and \textit{Conciseness}. THumB \cite{kasai2022transparent} contains 2,500 human judgments for 500 images across two dimensions: \textit{Precision} and \textit{Recall}. LongCap-Arena \cite{vela} provides 32,246 human judgments for 7,805 images across three dimensions: \textit{Descriptiveness}, \textit{Relevance}, and \textit{Fluency}. Despite these efforts, existing datasets are typically limited to either short or long captions and lack diverse image content as well as fine-grained MOS annotations. Moreover, these benchmarks do not evaluate the capability of MLLMs in image caption assessment.

\subsection{Image Caption Evaluation Metric}
Automatic caption evaluation has traditionally relied on reference-based lexical overlap metrics \cite{papineni2002bleu,banerjee2005meteor,lin2004rouge}. Later, consensus-oriented metrics are proposed by comparing generated captions with multiple reference captions \cite{cider,spice}. However, these classic metrics are shown to correlate weakly 
with human judgments \cite{clipscore,sarto2024bridge}. Recent work has explored reference-free evaluation using pretrained vision-language representations \cite{clipscore,radford2021clip,sarto2024bridge}, but these methods focus primarily on short captions and are limited to a single evaluation aspect. To better capture fine-grained visual details and richer semantic content, \cite{yao2024hifi,vela} propose evaluation methods designed for long captions, but they do not differentiate the evaluation dimensions for short and long captions. With the development of MLLMs, \cite{lee2024fleur,ohi2024multi,tong2025g} leverage them to provide more interpretable evaluations from multiple perspectives, but their performance remains insufficient in aligning with human preference \cite{vela}. Therefore, a unified automatic evaluation metric that is suitable for both short and long captions and aligns with human judgment is still required.

% \begin{table*}[t]
%   \caption{Representative benchmarks/protocols and metrics for image caption evaluation.
%   ``Long'' indicates whether long, detailed captions are a primary focus; ``Human'' indicates whether human judgments are used.
%   For metrics, ``Ref'' indicates whether reference captions are required.}
%   \label{tab:rw_bench_metric}
%   \centering
%   \small
%   \setlength{\tabcolsep}{6pt}
%   \renewcommand{\arraystretch}{1.1}

%   % Symbols (robust, no extra math mode issues)

%   \begin{tabular}{p{0.42\linewidth} p{0.16\linewidth} c c}
%     \toprule
%     \multicolumn{4}{l}{\textbf{Benchmarks / Protocols}}\\
%     \midrule
%     \textbf{Name} & \textbf{Scale} & \textbf{Long} & \textbf{Human} \\
%     \midrule
%     LongCap-Arena (VELA)~\cite{vela} & -- & \cmark & \cmark \\
%     Flickr8K-Expert / Flickr8K-CF & 8{,}000 imgs & \xmark & \cmark \\
%     Web-2040  & 2{,}040 imgs & \xmark & \xmark \\
%     \midrule
%     \multicolumn{4}{l}{\textbf{Metrics}}\\
%     \midrule
%     \textbf{Metric} & \textbf{Ref} & \textbf{Type} & \textbf{Measures} \\
%     \midrule
%     Descriptiveness  & \xmark & axis & detail completeness / coverage \\
%     Relevance  & \xmark & axis & correctness / grounding \\
%     CIDEr~\cite{cider} & \cmark & overlap & consensus n-gram similarity \\
%     SPICE~\cite{spice} & \cmark & semantic & proposition / scene-graph match \\
%     CLIPScore~\cite{clipscore} & \xmark & VL-score & image--text compatibility \\
%     CLIP i2i (extrinsic) & \xmark & VL-score & generated--original similarity \\
%     \bottomrule
%   \end{tabular}
% \end{table*}

\section{ICBench}
In this section, we introduce \textbf{ICBench}, the first large-scale image captioning dataset with fine-grained MOSs for both short and long captions. The dataset contains 2,040 high-quality images, along with 20,400 short captions and 20,400 long captions generated by 10 advanced MLLMs. It further includes 1.8M human annotations evaluating fluency, relevance, and conciseness for short captions, and fluency, relevance, and completeness for long captions. ICBench provides a comprehensive resource for image captioning evaluation and enables systematic benchmarking of the captioning capabilities of MLLMs.

\subsection{Data Collection}
To ensure diversity in image content, we first define 12 categories covering a wide range of real-world scenarios: portrait, architecture, landscape, food, art, fashion, drinks, animals, sports, technology, transportation, and indoor scenes. Images are collected from free photography websites, with a minimum resolution of 1024×1024 to guarantee visual quality. To make the benchmark comprehensive, most images contain multiple main objects, and all key targets are clearly visible to avoid ambiguity or confusing compositions. This ensures that the dataset reflects a variety of visual contexts and interactions, providing a robust testbed for evaluating image captioning performance.

Next, we select 10 advanced MLLMs to generate captions for the dataset, including eight open-source models—LLaVA-1.6 \cite{llava}, LLama3.2-V \cite{llama}, MiniCPM-V2.6 \cite{minicpm}, mPLUG-Owl3 \cite{mplug}, Ovis2.5 \cite{ovis25}, DeepSeekVL2 \cite{deepseekv2}, InternVL3.5 \cite{internvl3_5}, and Qwen3-VL \cite{qwen3}, as well as two closed-source models, ChatGPT-5 \cite{gpt5} and Gemini3 \cite{gemini3}. For each image, both short and long captions are generated. Short captions are designed to describe the most salient 1-2 objects or concepts in the image, providing a concise summary of the main content. In contrast, long captions aim to describe as much information as possible in the image, including all visible objects, attributes, and their relationships. This design ensures that the benchmark covers a wide range of caption lengths, from concise descriptions capturing key content to more elaborate captions that convey detailed context, relationships, and attributes. Totally, we have generated 20,400 short captions and 20,400 long captions.
\begin{figure*}
\centering
  \includegraphics[width=1\textwidth]{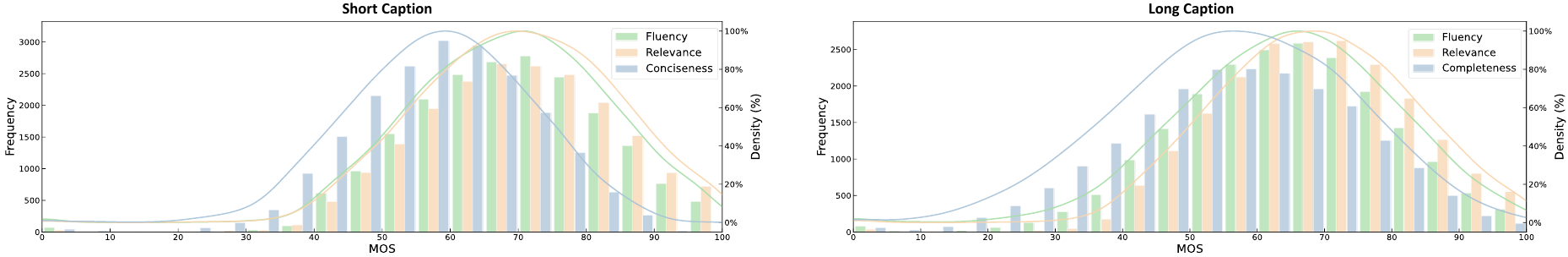}
  \caption{MOS distribution of short caption and long caption across different evaluation dimensions.}
  \label{mos}
\end{figure*}
\begin{figure*}
\centering
  \includegraphics[width=1\textwidth]{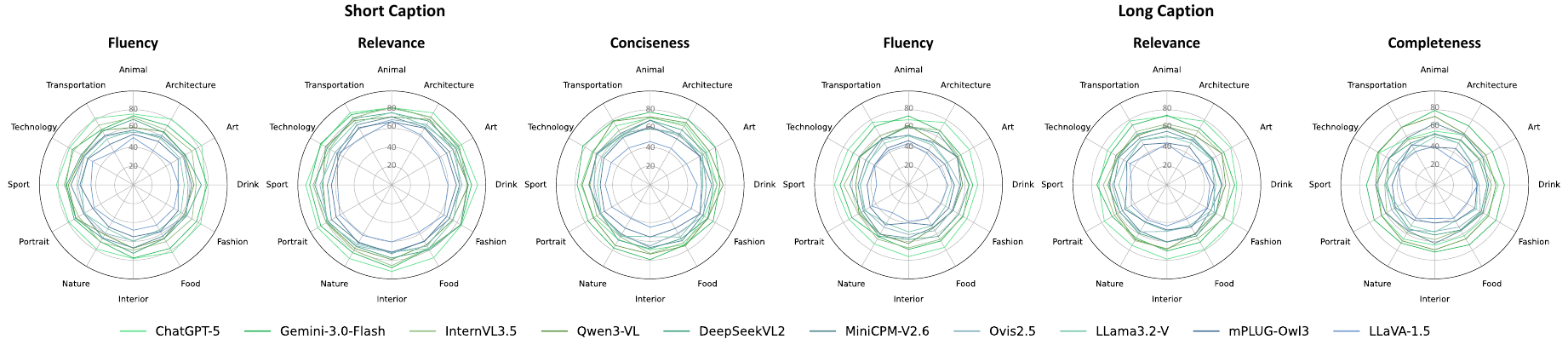}
  \caption{Performance comparison of different MLLMs on short captioning in terms of fluency, relevance, and conciseness, and on long captioning in terms of fluency, relevance, and completeness across different image contents.}
  \label{catcomparison}
\end{figure*}

\subsection{Subjective Experiment Setup}
To evaluate the edited images, we conduct a subjective quality assessment experiment using the ICBench. This experiment is designed to capture human judgment for caption, ensuring the results align with real-world human perception.

As shown in Figure~\ref{overall}(b), during the experiment, participants are presented with an image with its corresponding short caption and long caption. Participants are asked to assess the edited image using a 5-point continuous scale from three aspects for short captions:
\begin{itemize}[left=0pt, labelsep=0.6em, labelwidth=0pt]
    \item \textbf{Fluency:} The caption is grammatical correctness, coherence, and naturalness of the caption, without grammatical or spelling errors.
    \item \textbf{Relevance:} The caption describes the main content of the image, focusing on the most important objects and scene elements, and avoids including unrelated or irrelevant information.
    \item \textbf{Conciseness:} The caption is clear and succinct, without redundant wording.
\end{itemize}  
and three aspects for long captions:
\begin{itemize}[left=0pt, labelsep=0.6em, labelwidth=0pt]
    \item \textbf{Fluency:} The caption is grammatical correctness, coherence, and naturalness of the caption, without grammatical or spelling errors.
    \item \textbf{Relevance:} The caption describes the main content of the image, focusing on the most important objects and scene elements, and avoids including unrelated or irrelevant information.
    \item \textbf{Completeness:} The caption thoroughly covers all visible objects, attributes, and relationships present in the image.
\end{itemize}  

The experiment is conducted using a Python-based GUI displayed on a calibrated LED monitor with a resolution of 3840 $\times$ 2160, with images shown in 1024$\times$1024 resolution in a random order. A total of 30 professional annotators, seated 2 feet from the monitor in a controlled environment, complete the study in 45 sessions, each under 30 minutes, to mitigate fatigue. Each caption is assessed by 15 participants. All annotators undergo rigorous training following a standardized protocol. A pre-test is conducted to assess participants’ comprehension of the criteria and their alignment with the standard examples. Participants who did not meet the required accuracy threshold were excluded from further participation. In total, we collect 1,836K ratings (15 annotators $\times$ 40,800 captions $\times$ 3 dimensions).

\subsection{Subjective Data Processing}
We follow the guidelines outlined in \cite{subject,duan2022confusing} to identify and exclude outliers, as well as to reject subjects who provide unreliable ratings. An individual rating for an caption is considered an outlier if it falls outside $2$ standard deviations (if normal) or $\sqrt{20}$
standard deviations (if not normal) from the mean rating of that image \cite{subject}. A subject is excluded if over $5\%$ of their ratings are outliers. 
As a result, no subject is excluded based on this criterion and $1.83\%$ of the total subjective ratings are removed. 
The remaining valid ratings are converted into Z-scores \cite{subject}, then linearly scaled to the range [0,100]. The final MOS is calculated as follows
\begin{equation}
    z_{ij} = \frac{r_{ij} - \mu_i}{\sigma_i}, \ z_j = \frac{1}{N_j} \sum_{i=1}^{N_j} z_{ij}, \
    MOS_j = \frac{100(z_j + 3)}{6}
\end{equation}
where where $r_{ij}$ is the raw rating given by the i-th subject to the j-th caption, $\mu_i$ is the mean rating and $\sigma_i$ is the standard deviation provided by the i-th subject and $N_j$ is the number of valid ratings for the j-th caption.

\subsection{Subjective Data Analysis}

Figure~\ref{mos} shows the MOS distributions, providing an overview of the overall performance of all models across the three dimensions. We observe that fluency and relevance scores are similar for short captions, but show greater variation for long captions, likely because long captions require more detailed descriptions and reasoning, which introduces more room for errors in sentences. The conciseness and completeness dimensions exhibit lower MOS values, indicating that models struggle to generate succinct short captions and fully comprehensive long captions.
\begin{figure*}[t]
\centering
  \includegraphics[width=1\textwidth]{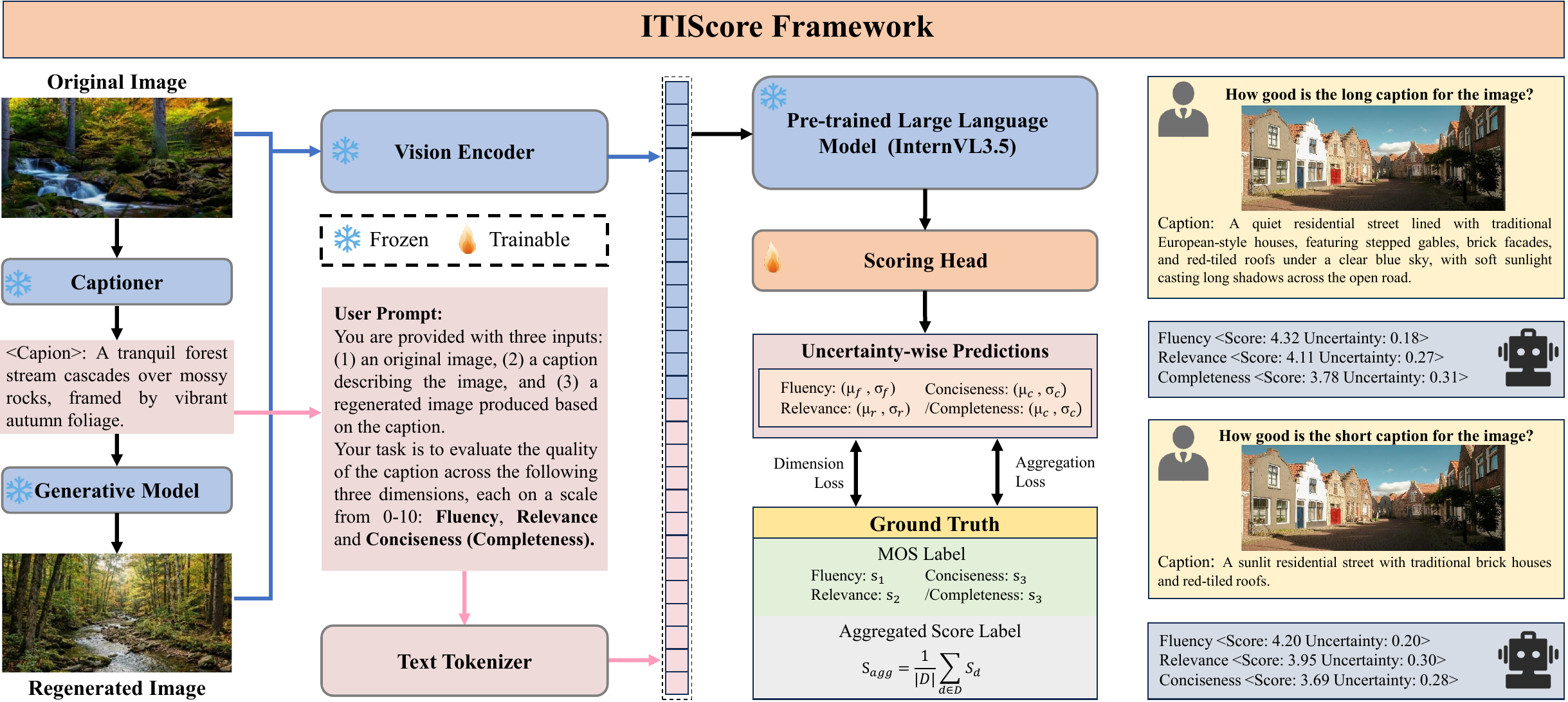}
  \caption{Overview of our ITIScore. Given an image and its caption, a pretrained generative model reconstructs an image from the caption. The original image, reconstructed image, and caption are jointly fed into a multimodal large language model to obtain a unified representation. A lightweight MLP scoring head then predicts the mean score and uncertainty for each evaluation dimension, enabling uncertainty-aware multi-dimensional caption quality assessment.}
  \label{model}
\end{figure*}
We compare the short- and long-captioning abilities of various MLLMs based on MOSs, as shown in Figure~\ref{catcomparison}. Closed-source models, including ChatGPT-5~\cite{gpt5} and Gemini-3.0-Flash~\cite{gemini3}, outperform open-source models overall. Among the latter, InternVL3.5~\cite{internvl3_5} and Qwen3-VL~\cite{qwen3} perform relatively well, while LLaVA-1.5 \cite{llava} performs the worst. In short captioning, ChatGPT-5 achieves high fluency and relevance but is less concise, scoring below Gemini-3.0-Flash and Qwen3-VL. Most models struggle with conciseness for food and art images due to their rich visual details. In long captioning, ChatGPT-5 remains strong in fluency and relevance but underperforms in completeness compared to Gemini-3.0-Flash. While fluency is generally high across categories, models show weaker relevance and completeness on architecture and technology images, likely due to the need for more specialized knowledge.

\section{ITIScore}
In this section, we introduce ITIScore, a unified evaluation framework for both short and long image captioning across multiple dimensions. The framework is built upon an image-to-text-to-image paradigm and leverages an MLLM to fuse visual and textual features, while keeping the MLLM backbone frozen.
% TODO: models, prompts, decoding, output format (jsonl), categories

\subsection{Architecture Design}
The overall architecture of ITIScore follows an image-to-text-to-image evaluation paradigm, which leverages both visual reconstruction and multimodal reasoning to assess caption quality. 
Given an input image $I$ and a generated caption $C$, the model first reconstructs an image from the caption and then evaluates the caption by comparing the original image, the reconstructed image, and the caption within a MLLM. This design enables the model to capture both low-level visual consistency and high-level semantic alignment between the caption and the image.

Formally, given an image-caption pair $(I, C)$, we employ a pretrained generative model to reconstruct a corresponding image $I'$ conditioned on the caption:
\begin{equation}
I' = G(C),
\end{equation}
where $G(\cdot)$ denotes the generative model. The reconstructed image $I'$ serves as a visual proxy to evaluate how well the caption captures the original visual content. By comparing $I'$ with $I$, the model can identify whether the caption faithfully describes salient objects, attributes, and relationships present in the original image.

To evaluate the overall caption quality, we jointly consider the original image $I$, the reconstructed image $I'$, and the caption $C$. These inputs are fed into a MLLM, which encodes visual and textual information into a unified latent representation through a projector. Let $f_{\text{MLLM}}(\cdot)$ denote the multimodal backbone. The hidden representation is obtained as
\begin{equation}
H = f_{\text{MLLM}}(I, I', C),
\end{equation}
where $H$ represents the final hidden states of the MLLM, containing rich cross-modal contextual information. By incorporating both the original and reconstructed images, the model can reason about inconsistencies or omissions in the caption that may not be apparent from the caption alone.

The final hidden states are passed to a lightweight scoring head $R_{\omega}$, which is implemented as a three-layer MLP and predicts the parameters of a Gaussian distribution:
\begin{equation}
(\mu, \sigma) = R_{\omega}(H),
\end{equation}
where $\mu$ denotes the predicted mean quality score and $\sigma$ captures the uncertainty associated with the prediction. Modeling the output as a Gaussian distribution allows the scoring head to express uncertainty for ambiguous captions or visually complex images. This uncertainty-aware formulation also enables the model to provide more conservative scores for low-confidence predictions while preserving high resolution for confident ones.
\begin{table*}[t]
\setlength{\tabcolsep}{8pt} 
\centering
\caption{The Kendall correlation coefficient $\tau_b$ and $\tau_c$ between human judgments and various automatic metrics on ICBench. The first part is for reference-free methods, and the second part is advanced MLLMs. The best results are highlighted in \mredbf{red}, and the second-best results are highlighted in \mbluebf{blue}.}
\resizebox{\linewidth}{!}{
\begin{tabular}{l||cccccc:cccccc:cc}
\toprule
Subset& \multicolumn{6}{c}{Short Caption} & \multicolumn{6}{c}{Long Caption}\\
\cmidrule(lr){2-7} \cmidrule(lr){8-13}
Dimension& \multicolumn{2}{c}{Fluency} & \multicolumn{2}{c}{Relevance}& \multicolumn{2}{c}{Conciseness}&\multicolumn{2}{c}{Fluency} & \multicolumn{2}{c}{Relevance}& \multicolumn{2}{c}{Completeness}&\multicolumn{2}{c}{Overall} \\
\cmidrule(lr){2-3} \cmidrule(lr){4-5} \cmidrule(lr){6-7} \cmidrule(lr){8-9} \cmidrule(lr){10-11} \cmidrule(lr){12-13} \cmidrule(lr){14-15}
Method/Metric&$\tau_b$ &$\tau_c$  & $\tau_b$ & $\tau_c$& $\tau_b$ & $\tau_c$&$\tau_b$ &$\tau_c$  & $\tau_b$ & $\tau_c$& $\tau_b$ & $\tau_c$ & $\tau_b$ & $\tau_c$\\
\midrule
UMIC~\cite{lee2021umic} & 32.17 & 31.94 & 18.43 & 17.21 & 6.88 & 5.67 & 29.76 & 28.62 & 7.12 & 6.97 & 8.83 & 7.69 &17.20&16.35 \\
CLIP-S~\cite{clipscore} & 35.42 & 35.31 & 19.81 & 18.73 & 7.91 & 6.87 & 30.61 & 29.53 & 10.02 & 9.91 & 9.79 & 8.72&18.93&18.17 \\
PAC-S~\cite{sarto2023positive} & 40.47 & 39.32 & 21.42 & 20.82 & 10.23 & 6.68 & 32.79 & 31.63 & 13.47 & 12.31 & 10.87 & 9.76&21.54&20.09 \\
FLEUR~\cite{lee2024fleur} & 36.18 & 35.77 & 22.69 & 21.56 & 6.13 & 5.03 & 31.53 & 30.42 & 11.83 & 10.72 & 8.12 & 7.01&19.41&18.42 \\
\midrule
LLaVA-1.5 (7B) \cite{llava} & 25.99 & 26.04 & 21.06 & 21.45 & 21.21 & 21.55 & 23.92 & 23.95 & 21.77 & 22.02 & 24.75 & 24.85 & 23.12 & 23.31 \\
mPLUG-Owl3 (7B) \cite{mplug} & 30.47 & 30.30 & 33.92 & 33.72 & 26.68 & 26.72 & 29.46 & 29.25 & 27.82 & 27.86 & 28.46 & 28.25 & 29.47 & 29.35 \\
MiniCPM-V2.6 (8B) \cite{minicpm}& 28.53 & 27.08 & 24.42 & 23.14 & 24.66 & 23.46 & 27.11 & 25.83 & 22.99 & 21.94 & 25.43 & 24.17 & 25.52 & 24.27 \\
Ovis2.5 (8B) \cite{ovis25} & 25.59 & 24.43 & 18.69 & 18.06 & 17.81 & 17.26 & 25.16 & 24.07 & 16.64 & 16.14 & 23.81 & 22.80 & 21.28 & 20.46 \\
LLama3.2-V (11B) \cite{llama} & 40.66 & 40.27 & 39.47 & 39.24 & 36.06 & 36.01 & 38.48 & 38.31 & 34.57 & 34.47 & 37.88 & 37.57 & 37.85 & 37.65 \\
DeepSeekVL2 (small) \cite{deepseekv2} & 39.74 & 39.44 & 32.83 & 32.70 & 33.34 & 33.35 & 38.74 & 38.44 & 32.17 & 32.25 & 36.71 & 36.42 & 35.59 & 35.43 \\
Qwen2.5-VL (8B) \cite{qwenvl2} & 44.90 & 44.48 & 33.91 & 33.79 & 29.19 & 29.16 & 43.79 & 43.28 & 31.05 & 31.03 & 43.52 & 43.02 & 37.73 & 37.46 \\
Qwen3-VL (8B) \cite{qwen3} & 48.43 & 47.71 & 41.25 & 40.88 & 37.96 & 37.79 & 49.54 & 48.91 & 44.60 & 44.34 & 44.44 & 43.89 & 44.37 & 43.92 \\
InternVL3 (8B) \cite{internvl3} & 42.58 & 42.08 & 29.07 & 29.16 & 26.22 & 26.29 & 41.56 & 41.04 & 29.86 & 29.73 & 38.83 & 38.56 & 34.69 & 34.48 \\
InternVL3.5 (8B) \cite{internvl3_5} & 47.57 & 46.83 & 44.38 & 43.83 & 37.96 & 37.82 & 46.73 & 46.18 & 40.49 & 40.35 & 43.24 & 42.86 & 43.40 & 42.98 \\
Gemini-3.0-Flash \cite{gemini3} & 49.07 & 48.48 & 48.98 & 48.28 & 37.86 & 37.62 & 51.02 & 50.20 & 43.21 & 42.77 & 46.52 & 46.07 & 46.11 & 45.57 \\
ChatGPT-5 \cite{gpt5} & 52.23 & 51.37 & 46.80 & 46.13 & 42.59 & 42.36 & 49.81 & 49.17 & 45.36 & 45.11 & 45.76 & 45.00 & 47.09 & 46.52 \\
\midrule
ITIScore (DeepSeekVL2 (small)) & 62.40 & 62.40 & \mbluebf{61.24} & \mbluebf{61.25} & 57.79 & 57.79 & \mbluebf{64.45} & \mbluebf{64.45} & 56.02 & 56.02 & 58.48 & 58.48 & 60.06 & 60.06 \\
ITIScore (Qwen3-VL (8B)) & \mbluebf{64.04} & \mbluebf{64.04} & 59.19 & 59.20 & \mbluebf{60.54} & \mbluebf{60.54} & 63.84 & 63.85 & \mbluebf{57.24} & \mbluebf{57.24} & \mbluebf{59.08} & \mbluebf{59.08} & \mbluebf{60.66} & \mbluebf{60.66} \\
 \rowcolor{gray!20}  % 这里添加灰色阴影
ITIScore (InternVL3.5 (8B)) & \mredbf{64.48} & \mredbf{64.49} & \mredbf{63.21} & \mredbf{63.21} & \mredbf{62.10} & \mredbf{62.10} & \mredbf{65.39} & \mredbf{65.39} & \mredbf{60.65} & \mredbf{60.65} & \mredbf{61.71} & \mredbf{61.72} & \mredbf{62.92} & \mredbf{62.93} \\
\bottomrule
\end{tabular}
}
\label{single}
\end{table*}
This architecture design allows ITIScore to evaluate caption quality in a robust and interpretable manner. By integrating visual reconstruction consistency with multimodal semantic understanding, the model can assess whether a caption is not only semantically faithful to the original image but also coherent and informative. Moreover, the use of a Gaussian distribution for output further supports uncertainty-aware evaluation, which is crucial for aligning automated scoring with human subjective judgments.
% Instead of directly using the mean score, we define the final quality score as the expected probability that the latent quality exceeds a satisfaction threshold $c$, which is set as the middle point of the score level. The predicted quality score $S$ can be expressed as:
% \begin{equation}
% S=
% \int_{-\infty}^{\infty}
% \sigma(r-c)\,
% \mathcal{N}(r|\mu(x),\sigma(x))\,dr,
% \end{equation}
% where $\sigma(\cdot)$ denotes the sigmoid function. This formulation naturally incorporates both the predicted mean and uncertainty. However, the above logistic--normal integral does not admit a closed-form solution. Following the commonly used logistic-normal approximation, the expectation can be approximated analytically as
% \begin{equation}
% S
% \approx
% \sigma
% \left(
% \frac{\mu-c}{\sqrt{1+\frac{\pi}{8}\sigma^2}}
% \right).
% \end{equation} 
% When the uncertainty increases, the denominator becomes larger, leading to a more conservative score, reflecting lower confidence in the prediction. 

\subsection{Training Strategy}
Inspired by \cite{duan2025finevq,wang2026quality,wang2025lmm4lmm}, we utilize the instruction tuning to support multi-dimensional caption evaluation. The instruction specifies the evaluation dimension, such as \textit{Fluency}, \textit{relevance}, \textit{conciseness}, or \textit{completeness}. By conditioning the scoring head on the instruction, the model is able to predict distribution parameters $(\mu, \sigma)$ corresponding to the requested dimension. This design allows a single model to flexibly handle multiple evaluation criteria.

Given the ground-truth MOSs for each dimension, denoted by $s_d$ for $d \in \mathcal{D}$, we train the model using a dimension-wise Gaussian negative log-likelihood (NLL) objective:
\begin{equation}
\mathcal{L}_{\text{dim}} =
\frac{1}{|\mathcal{D}|} \sum_{d \in \mathcal{D}} 
\left(
\frac{(s_d - \mu_d)^2}{2\sigma_d^2} + \frac{1}{2} \log \sigma_d^2
\right) .
\end{equation}
where $\mathcal{D} = \{\textit{Fluency}, \textit{relevance}, \textit{conciseness}\}$ for short captions, and 
$\mathcal{D} = \{\textit{Fluency}, \textit{relevance}, \textit{completeness}\}$ for long captions. This objective encourages the predicted mean $\mu_d$ for each dimension to approximate the corresponding MOS while also predicting an uncertainty $\sigma_d$ that captures the ambiguity of that dimension. A larger $\sigma_d$ reduces the contribution of ambiguous or difficult samples to the loss, while a smaller $\sigma_d$ encourages the model to closely match the ground-truth MOS for straightforward samples.

We further define an aggregated score to represent overall quality across all evaluated dimensions. Let $\mu_d$ denote the predicted mean score for each dimension $d$. The aggregated score is computed as
\begin{equation}
\mu_{\text{agg}} =
\frac{1}{|\mathcal{D}|}
\sum_{d \in \mathcal{D}}
\mu_d ,
\end{equation}

Because the sum of Gaussian random variables is also Gaussian, we can model the aggregated score distribution as
\begin{equation}
s_{\text{agg}}
\sim
\mathcal{N}
\left(
\sum_d \mu_d,
\sum_d \sigma_d^2
\right),
\end{equation}
which allows us to capture the overall uncertainty of the combined multi-dimensional score. To ensure that the aggregated score aligns with the overall perceived quality, we introduce an additional regression loss:
\begin{equation}
\mathcal{L}_{agg}
=
\left\|
\mu_{\text{agg}} - s_{\text{agg}}
\right\|^2 .
\end{equation}
This constraint encourages consistency between the individual dimension predictions and the overall quality assessment, thereby promoting coherent multi-dimensional evaluation.

The final training objective combines the uncertainty-aware regression loss for individual dimensions and the aggregated score constraint:
\begin{equation}
\mathcal{L}
=
\mathcal{L}_{dim}
+
\lambda
\mathcal{L}_{agg}.
\end{equation}
During training, the parameters of the MLLM backbone are frozen to preserve pretrained knowledge, while only the scoring head is optimized. This design enables the model to efficiently learn to predict both mean scores and associated uncertainties, while simultaneously maintaining consistency across multiple evaluation dimensions.
\begin{table*}[t]
\belowrulesep=0pt
\aboverulesep=0pt
\centering

\renewcommand{\arraystretch}{0.85}
\caption{Comparisons of the alignment between different evaluation methods and human perception in short captions. The best results are highlighted in \mredbf{red}, and the second-best results are highlighted in \mbluebf{blue}. The Spearman Rank Correlation Coefficient (SRCC) is reported.} 
\resizebox{\textwidth}{!}{
\begin{tabular}{l||c:ccc| c:ccc| c:ccc| c:ccc| c:c |c:c}
\toprule
\noalign{\vspace{1.5pt}}
Dimensions &  \multicolumn{4}{c}{Fluency} & \multicolumn{4}{c}{Relevance}&\multicolumn{4}{c}{Conciseness}& \multicolumn{2}{c}{Overall Rank}\\
\cmidrule(lr){2-5}
\cmidrule(lr){6-9}
\cmidrule(lr){10-13}
\cmidrule(lr){18-19}
\noalign{\vspace{1.5pt}}
Models/Metrics & Human &\cellcolor{gray!20} Ours&GPT-5 \cite{gpt5}&FLEUR \cite{lee2024fleur}&Human&\cellcolor{gray!20}Ours&GPT-5 \cite{gpt5}&FLEUR \cite{lee2024fleur}&Human&\cellcolor{gray!20}Ours&GPT-5 \cite{gpt5}&FLEUR \cite{lee2024fleur}&Human&\cellcolor{gray!20}Ours\\

\hline
\noalign{\vspace{1.5pt}}
ChatGPT-5 \cite{gpt5} & 80.13 & \cellcolor{gray!20}63.37 & 57.76 & 69.12 & 88.55 & \cellcolor{gray!20}62.32 & 61.29 & 69.80 & 70.78 & \cellcolor{gray!20}59.84 & 49.78 & 57.30 & 1& \cellcolor{gray!20}1\\
Gemini-3.0-Flash \cite{gemini3} & 74.36 & \cellcolor{gray!20}60.94 & 55.49 & 67.20 & 84.57 & \cellcolor{gray!20}60.53 & 58.25 & 67.17 & 77.97 & \cellcolor{gray!20}58.72 & 54.41 & 61.41 & 2& \cellcolor{gray!20}2\\
InternVL3.5 \cite{internvl3_5} & 70.67 & \cellcolor{gray!20}60.04 & 52.88 & 65.69 & 81.81 & \cellcolor{gray!20}58.99 & 50.38 & 60.03 & 68.52 & \cellcolor{gray!20}56.71 & 45.34 & 56.11 & 3& \cellcolor{gray!20}3\\
Qwen3-VL \cite{qwen3} & 65.04 & \cellcolor{gray!20}59.17 & 51.45 & 61.92 & 77.42 & \cellcolor{gray!20}56.90 & 52.34 & 64.91 & 73.60 & \cellcolor{gray!20}57.46 & 51.42 & 54.70 & 4& \cellcolor{gray!20}4\\
DeepSeekVL2 \cite{deepseekv2} & 67.69 & \cellcolor{gray!20}58.25 & 54.06 & 63.96 & 79.52 & \cellcolor{gray!20}58.70 & 54.67 & 63.94 & 66.44 & \cellcolor{gray!20}55.49 & 46.65 & 54.38 & 5& \cellcolor{gray!20}5\\
MiniCPM-V2.6 \cite{minicpm} & 62.42 & \cellcolor{gray!20}57.92 & 47.12 & 64.73 & 71.13 & \cellcolor{gray!20}54.20 & 47.79 & 63.18 & 64.46 & \cellcolor{gray!20}52.79 & 47.97 & 53.67 & 6& \cellcolor{gray!20}6\\
Ovis2.5 \cite{ovis25} & 59.84 & \cellcolor{gray!20}56.22 & 48.90 & 59.84 & 73.34 & \cellcolor{gray!20}55.81 & 49.60 & 61.63 & 62.42 & \cellcolor{gray!20}53.83 & 43.64 & 57.34 & 7&\cellcolor{gray!20} 7\\
LLama3.2-V \cite{llama} & 56.91 & \cellcolor{gray!20}55.33 & 50.41 & 62.83 & 75.37 & \cellcolor{gray!20}56.89 & 47.51 & 62.61 & 60.07 & \cellcolor{gray!20}54.98 & 39.45 & 50.76 & 8&\cellcolor{gray!20} 8\\
mPLUG-Owl3 \cite{mplug} & 53.46 & \cellcolor{gray!20}54.09 & 45.51 & 58.23 & 68.33 & \cellcolor{gray!20}53.48 & 45.74 & 58.39 & 56.91 & \cellcolor{gray!20}51.92 & 41.96 & 48.95 & 9&\cellcolor{gray!20} 9\\
LLaVA-1.5 \cite{llava} & 46.94 & \cellcolor{gray!20}51.72 & 42.24 & 53.77 & 61.71 & \cellcolor{gray!20}51.36 & 42.11 & 53.42 & 46.36 & \cellcolor{gray!20}49.18 & 34.80 & 55.28 & 10&\cellcolor{gray!20} 10\\
\midrule
SRCC to human$\uparrow$ &  & \cellcolor{gray!20}\mredbf{0.983} & \mbluebf{0.917} & 0.883 &  & \cellcolor{gray!20}\mredbf{1.000} & \mbluebf{0.900} & 0.683 &  & \cellcolor{gray!20}\mredbf{0.933} & \mbluebf{0.917} & 0.500 &  &\cellcolor{gray!20}\mredbf{1.000} \\
\bottomrule
\end{tabular}
}
\label{compare_llm_s}
\end{table*}

\begin{table*}[t]
\belowrulesep=0pt
\aboverulesep=0pt
\centering
\renewcommand{\arraystretch}{0.85}
\caption{Comparisons of the alignment between different evaluation methods and human perception in long captions. The best results are highlighted in \mredbf{red}, and the second-best results are highlighted in \mbluebf{blue}. The Spearman Rank Correlation Coefficient (SRCC) is reported.} 
\resizebox{\textwidth}{!}{
\begin{tabular}{l||c:ccc| c:ccc| c:ccc| c:ccc| c:c |c:c}
\toprule
\noalign{\vspace{1.5pt}}
Dimensions &  \multicolumn{4}{c}{Fluency} & \multicolumn{4}{c}{Relevance}&\multicolumn{4}{c}{Completeness}& \multicolumn{2}{c}{Overall Rank}\\
\cmidrule(lr){2-5}
\cmidrule(lr){6-9}
\cmidrule(lr){10-13}
\cmidrule(lr){18-19}
\noalign{\vspace{1.5pt}}
Models/Metrics & Human &\cellcolor{gray!20} Ours&GPT-5 \cite{gpt5}&FLEUR \cite{lee2024fleur}&Human&\cellcolor{gray!20}Ours&GPT-5 \cite{gpt5}&FLEUR \cite{lee2024fleur}&Human&\cellcolor{gray!20}Ours&GPT-5 \cite{gpt5}&FLEUR \cite{lee2024fleur}&Human&\cellcolor{gray!20}Ours\\

\hline
\noalign{\vspace{1.5pt}}
ChatGPT-5 \cite{gpt5} & 74.62 & \cellcolor{gray!20}50.73 & 51.72 & 64.54 & 76.57 & \cellcolor{gray!20}50.44 & 63.53 & 52.74 & 62.83 & \cellcolor{gray!20}47.88 & 56.31 & 47.69 & 1& \cellcolor{gray!20}1 \\
Gemini-3.0-Flash \cite{gemini3} & 68.53 & \cellcolor{gray!20}48.44 & 52.97 & 61.07 & 70.40 & \cellcolor{gray!20}48.96 & 58.36 & 51.86 & 73.00 & \cellcolor{gray!20}49.37 & 62.27 & 45.15 & 2& \cellcolor{gray!20}2 \\
InternVL3.5 \cite{internvl3_5} & 64.45 & \cellcolor{gray!20}47.92 & 55.66 & 59.00 & 66.31 & \cellcolor{gray!20}47.47 & 60.13 & 53.46 & 56.64 & \cellcolor{gray!20}44.38 & 52.55 & 45.36 & 3& \cellcolor{gray!20}4 \\
Qwen3-VL \cite{qwen3} & 58.21 & \cellcolor{gray!20}46.83 & 49.56 & 56.79 & 62.90 & \cellcolor{gray!20}46.44 & 56.73 & 44.50 & 66.90 & \cellcolor{gray!20}46.12 & 59.10 & 43.68 & 4& \cellcolor{gray!20}3\\
DeepSeekVL2 \cite{deepseekv2} & 61.16 & \cellcolor{gray!20}45.70 & 50.75 & 51.05 & 59.98 & \cellcolor{gray!20}43.03 & 51.05 & 49.66 & 53.91 & \cellcolor{gray!20}42.57 & 54.70 & 44.02 & 5& \cellcolor{gray!20}5\\
MiniCPM-V2.6 \cite{minicpm} & 55.35 & \cellcolor{gray!20}44.60 & 45.56 & 53.04 & 57.20 & \cellcolor{gray!20}44.17 & 52.99 & 43.45 & 59.58 & \cellcolor{gray!20}45.18 & 50.53 & 46.60 & 6& \cellcolor{gray!20}6\\
Ovis2.5 \cite{ovis25} & 52.55 & \cellcolor{gray!20}43.32 & 46.76 & 55.21 & 54.36 & \cellcolor{gray!20}45.37 & 54.76 & 52.75 & 51.06 & \cellcolor{gray!20}41.85 & 48.81 & 42.66 & 7& \cellcolor{gray!20}7\\
LLama3.2-V \cite{llama} & 49.50 & \cellcolor{gray!20}42.78 & 48.44 & 45.98 & 51.22 & \cellcolor{gray!20}42.23 & 48.85 & 46.09 & 48.14 &\cellcolor{gray!20}39.47 & 46.26 & 41.20 & 8& \cellcolor{gray!20}8\\
mPLUG-Owl3 \cite{mplug} & 45.84 & \cellcolor{gray!20}40.89 & 43.61 & 48.73 & 47.84 & \cellcolor{gray!20}40.85 & 46.59 & 47.35 & 44.73 & \cellcolor{gray!20}40.19 & 44.18 & 43.90 & 9& \cellcolor{gray!20}9\\
LLaVA-1.5 \cite{llava} & 40.01 & \cellcolor{gray!20}38.01 & 40.40 & 40.82 & 42.60 & \cellcolor{gray!20}38.77 & 42.60 & 41.92 & 39.54 & \cellcolor{gray!20}36.98 & 39.55 & 36.63 & 10& \cellcolor{gray!20}10\\
\midrule
SRCC to human$\uparrow$ &  & \cellcolor{gray!20}\mredbf{0.983} &\mbluebf{0.917} & 0.867 &  & \cellcolor{gray!20}\mredbf{0.933} & \mbluebf{0.917} & 0.533 &  & \cellcolor{gray!20}\mredbf{0.983} & \mbluebf{0.933} & 0.667 &  & \cellcolor{gray!20}\mredbf{0.983} \\
\bottomrule
\end{tabular}
}
\label{compare_llm_l}
\end{table*}

\section{Experiments}
In this section, we evaluate the performance of our ITIScore through extensive experiments.

\subsection{Experiment Setup}
We have split our ICBench to 4:1:1 for training, validating and testing. We respectively select InternVL3.5 (8B) \cite{internvl3_5}, DeepSeekVL2 (small), and Qwen3-VL (8B) \cite{qwen3} as the MLLM backbones, and use Qwen-Image \cite{qwenedit} as the generative model in our ITIScore framework. The models are implemented with PyTorch and trained on a 40GB NVIDIA RTX A100 GPU with a batch size of 1 and gradient accumulation steps of 16. The initial learning rate is set to 1e-4 and is decreased using the cosine annealing strategy.

\subsection{Evaluation on ICBench}
We compare ITIScore with existing caption evaluation methods on ICBench, including reference-free metrics and advanced MLLMs. Table~\ref{single} reports Kendall correlations with human judgments. Traditional reference-free metrics show low correlations, especially for semantic dimensions like relevance and completeness, indicating limited ability to capture fine-grained image–caption alignment. In contrast, MLLMs, including open-source models (Qwen3-VL, InternVL3.5) and closed-source models (Gemini-3.0-Flash, ChatGPT-5), achieve higher correlations, reflecting their strong multimodal understanding. It also shows that our benchmark can evaluate the captioning capabilities of MLLMs, showcasing their performance in both understanding and generating textual descriptions from visual content. Our ITIScore consistently outperforms all baselines across both short and long caption settings. All three variants of ITIScore significantly outperform existing metrics, demonstrating the robustness of the proposed framework. With InternVL3.5 (8B) as the backbone, it achieves the highest correlations on all six dimensions. These results demonstrate the effectiveness of the proposed image-to-text-to-image evaluation framework. Moreover, stronger MLLM backbones generally lead to better performance, with the InternVL3.5 backbone achieving the best overall results.

We further compare the alignment between different evaluation methods and human annotations in assessing the captioning performance of MLLMs, as presented in Table~\ref{compare_llm_s} and Table~\ref{compare_llm_l}. The overall ranking is determined by aggregating the average scores across the three evaluation dimensions. Compared with existing evaluation methods, such as ChatGPT-5 and FLEUR, our approach consistently yields rankings that better align with human preference. In the short-caption setting, ITIScore achieves the highest SRCC with human rankings across all dimensions, reaching 0.983, 1.000, and 0.933 for fluency, relevance, and conciseness, respectively, as well as 1.000 for the overall ranking. Similarly, in the long-caption setting, our method maintains the strongest alignment with human perception, obtaining SRCC scores of 0.983, 0.933, and 0.983 across the three dimensions and 0.983 for the overall ranking. 
\begin{table*}[tb]
\centering
\caption{Ablation study on our ICBench with different backbones and training strategies. The Kendall correlation coefficient $\tau_b$ with human judgments is reported.}
\label{ablation}
\setlength{\tabcolsep}{6pt} 
 \resizebox{1\textwidth}{!}{
\begin{tabular}{lcccc|ccc|ccc}
\toprule
\multicolumn{5}{c}{Backbone$\&$Strategy} & \multicolumn{3}{c}{Short Caption} & \multicolumn{3}{c}{Long Caption}\\
\cmidrule(lr){1-5} \cmidrule(lr){6-8} \cmidrule(lr){9-11} 
MLLM&Generation Model &  Scoring Head & $\mathcal{L}_{dim}$& $\mathcal{L}_{agg}$& Fluency&Relevance&Conciseness& Fluency&Relevance&Completeness\\ 
\hline
\noalign{\vspace{1pt}}
InternVL3.5 (8B) \cite{internvl3_5} &  &    & &  & 47.57  & 44.38 & 37.96 & 46.73  & 40.49  & 43.24 \\
InternVL3.5 (8B) \cite{internvl3_5} & Qwen-Image \cite{qwenedit} &    & &  & 47.55 & 51.60 & 42.06 & 47.20 & 50.71 & 45.28 \\
InternVL3.5 (8B) \cite{internvl3_5} &  & \checkmark  & Guassian NLL & \checkmark& 60.05 & 56.32 & 58.04 & 61.26 & 54.16 & 56.20 \\
InternVL3.5 (8B) \cite{internvl3_5} & Qwen-Image \cite{qwenedit} & \checkmark  & MSE& \checkmark & 59.57 & 56.55 & 57.28 & 60.84 & 55.82 & 56.50 \\
InternVL3.5 (8B) \cite{internvl3_5} & Qwen-Image \cite{qwenedit} & \checkmark  & L1& \checkmark & 58.84 & 56.46 & 56.70 & 60.00 & 55.52 & 56.45 \\
InternVL3.5 (8B) \cite{internvl3_5} & Qwen-Image \cite{qwenedit}  & \checkmark& Guassian NLL&  & 61.85 & 58.52 & 59.04 & 62.37 & 57.51 & 58.40 \\
\rowcolor{gray!20}
InternVL3.5 (8B) \cite{internvl3_5} & Qwen-Image \cite{qwenedit} & \checkmark  & Guassian NLL& \checkmark & 64.48 & \textbf{63.21} & 62.10 & \textbf{65.39} & \textbf{60.65} & 61.71 \\
InternVL3 (8B) \cite{internvl3} & Qwen-Image \cite{qwenedit} & \checkmark  & Guassian NLL& \checkmark & 61.00 & 57.52 & 58.40 & 62.08 & 56.25 & 57.46 \\
InternVL3.5 (14B) \cite{internvl3_5} & Qwen-Image \cite{qwenedit} & \checkmark  & Guassian NLL& \checkmark & \textbf{65.04} & 62.80 & \textbf{63.55} & 65.08 & 60.51 & \textbf{62.43} \\
InternVL3.5 (4B) \cite{internvl3_5} & Qwen-Image \cite{qwenedit} & \checkmark  & Guassian NLL& \checkmark & 58.52 & 55.15 & 56.72 & 59.50 & 54.36 & 55.24 \\
DeepSeekVL2 (small) \cite{deepseekv2} & Qwen-Image \cite{qwenedit} & \checkmark  & Guassian NLL& \checkmark & 62.40 & 61.24 & 57.79 & 64.45 & 56.02 & 58.48 \\
Qwen3-VL (8B) \cite{qwen3} & Qwen-Image \cite{qwenedit} & \checkmark  & Guassian NLL& \checkmark & 64.04 & 59.19 & 60.54 & 63.84 & 57.24 & 59.08 \\
InternVL3.5 (8B) \cite{internvl3_5} & FLUX.1 \cite{FLUX} & \checkmark & Guassian NLL & \checkmark & 61.22 & 58.41 & 58.55 & 62.24 & 56.50 & 57.05 \\
InternVL3.5 (8B) \cite{internvl3_5} & Stable Diffusion 3 \cite{SD} & \checkmark  & Guassian NLL & \checkmark& 61.10 & 57.55 & 58.15 & 61.58 & 56.02 & 57.94 \\
\bottomrule
\end{tabular}}
\end{table*}

\begin{table}[t]
\centering
\caption{Comparison of zero-shot results for Kendall correlation coefficient $\tau_b$ and $\tau_c$ between human judgments and various automatic metrics on Composites \cite{Composite}, Flickr8k-CF~\cite{Flickr8K-CF}, Flickr8k-Expert~\cite{Flickr8K-CF} datasets that have single evaluation dimension. The first part is for reference-based methods, the second part is for reference-free methods, and the third part is advanced MLLMs. The best results are highlighted in \mredbf{red}, and the second-best results are highlighted in \mbluebf{blue}.}
\resizebox{\linewidth}{!}{
\begin{tabular}{l||cccccc}
\toprule
 Dataset& \multicolumn{2}{c}{Composites \cite{Composite}} & \multicolumn{2}{c}{Flickr8k-CF \cite{Flickr8K-CF}}& \multicolumn{2}{c}{Flickr8k-Expert \cite{Flickr8K-CF}}\\
\cmidrule(lr){2-3} \cmidrule(lr){4-5} \cmidrule(lr){6-7}
Method/Metric&$\tau_b$ &$\tau_c$  & $\tau_b$ & $\tau_c$& $\tau_b$ & $\tau_c$ \\
 \noalign{\vspace{-1.5pt}}
\midrule
 BLEU~\cite{papineni2002bleu} & 29.0 & 31.3 & 32.2 & 32.3 & 17.9 & 9.3 \\
 ROUGE~\cite{lin2004rouge} & 30.0 & 32.4 & 32.1 & 32.3 & 19.9 & 10.3 \\
 METEOR~\cite{banerjee2005meteor} & 36.0 & 38.9 & 41.5 & 41.8 & 22.2 & 11.5 \\
 CIDEr~\cite{cider}& 34.9 & 37.7 & 43.6 & 43.9 & 24.6 & 12.7 \\
 SPICE~\cite{spice}& 38.8 & 40.3 & 51.7 & 44.9 & 24.4 & 12.0 \\
 BERT-S~\cite{zhang2020bertscore} & 28.4 & 30.1 & 38.5 & 39.2 & 22.8 & 15.7 \\
 MID~\cite{kim2022mutual} & 53.9 & 55.7 & 53.5 & 54.9 & 37.3 & 36.2 \\
 \noalign{\vspace{-1.5pt}}
\midrule
 UMIC~\cite{lee2021umic}& \mbluebf{55.3} & \mbluebf{56.1} & 47.3 & 46.8 & 34.1 & 35.7 \\
 CLIP-S~\cite{clipscore}& 49.8 & 53.8 &  51.1 & 51.2 & 34.4 & 17.7 \\
 PAC-S~\cite{sarto2023positive} & 51.5 & 55.7 & \mbluebf{53.9} & 54.3 & 36.0 & 18.6 \\
  \noalign{\vspace{-1.5pt}}
 \midrule
Qwen3-VL (7B) \cite{qwen3} & 37.2 & 34.5 & 39.1 & 40.0 & 29.1 & 20.2 \\
InternVL3.5 (7B) \cite{internvl3_5} & 41.0 & 38.3 & 42.8 & 44.1 & 31.5 & 22.5 \\
ChatGPT-4o \cite{chatgpt4o} & 46.2 & 42.7 & 48.0 & 49.5 & 36.8 & 25.0 \\
ChatGPT-5 \cite{gpt5} & 51.0 & 47.5 & 53.0 & \mbluebf{55.8} & 38.5 & \mbluebf{27.8} \\
Gemini-2.5-Flash \cite{nanobanana} & 44.0 & 43.5 & 45.5 & 46.8 & 35.0 & 24.2 \\
Gemini-3.0-Flash \cite{gemini3} & 49.2 & 49.8 & 50.8 & 52.5 & \mbluebf{39.8} & 26.5 \\
 \noalign{\vspace{-1.5pt}}
 \midrule
 \rowcolor{gray!20}  % 这里添加灰色阴影
 ITIScore (Ours)&\mredbf{55.2}&\mredbf{59.2}&\mredbf{56.6}&\mredbf{57.2}&\mredbf{41.8}&\mredbf{29.6}\\
  \noalign{\vspace{-1.5pt}}
\bottomrule
\end{tabular}
}
\label{single}
\end{table}

\begin{table}[t]
\centering
\setlength{\tabcolsep}{8pt} 
\caption{Comparison of zero-shot results for Kendall correlation coefficient $\tau_b$ between
human judgments and various automatic metrics on MMSE \cite{ohi2024multi} and LongCap-Arena~\cite{vela} datasets that have multiple evaluation dimensions. The first part is for reference-based methods, the second part is for reference-free methods, and the third part is advanced MLLMs. Flu: Fluency, Com: Completeness, Con: Conciseness, Desc: Descriptiveness, Rel: Relevance. The best results are highlighted in \mredbf{red}, and the second-best results are highlighted in \mbluebf{blue}.}
\resizebox{\linewidth}{!}{
\begin{tabular}{l||cccccc}
\toprule
Dataset & \multicolumn{3}{c}{MMHE \cite{ohi2024multi}}& \multicolumn{3}{c}{LongCap-Arena \cite{vela}} \\
\cmidrule(lr){2-4} \cmidrule(lr){5-7}
Method/Dimension&Flu.&Com.&Con.&Flu. & Desc.& Rel. \\
 \noalign{\vspace{-1.5pt}}
\midrule
 BLEU~\cite{papineni2002bleu}&8.8&8.7&10.7&14.5 & 30.3&6.3\\
 ROUGE~\cite{lin2004rouge}&8.5&7.4&11.6&6.2&8.3&5.7 \\
 METEOR~\cite{banerjee2005meteor} &9.1&9.8&6.6&4.6&11.2&8.1  \\
 CIDEr~\cite{cider}&14.4&9.8&9.6&3.2&5.5&5.1  \\
 BERT-S~\cite{zhang2020bertscore} &11.2&12.4&13.2&7.3&18.6&8.4  \\
\midrule
 CLIP-S~\cite{clipscore}& 6.2&13.4&5.4&25.0 &25.9&20.6 \\
 PAC-S~\cite{sarto2023positive} &26.4&27.3&10.6&23.3 & 26.2 & 21.7 \\
FLEUR~\cite{lee2024fleur}& 27.1&35.3&24.5&11.8&15.0&6.6 \\
 \noalign{\vspace{-1.5pt}}
 \midrule
Qwen3-VL (7B) \cite{qwen3}&36.8&33.5&38.2&30.7 &40.1&29.7\\
InternVL3.5 (7B) \cite{internvl3_5}&39.7&36.2&41.5&28.6 &43.0&31.8\\
ChatGPT-4o \cite{chatgpt4o}&45.2&41.5&47.2&23.1 &48.9&37.1\\
ChatGPT-5 \cite{gpt5}&45.6&\mbluebf{47.3}&\mbluebf{52.4}& \mbluebf{38.6}&50.1&\mbluebf{42.5}\\
Gemini-2.5-Flash \cite{nanobanana}&43.1&39.4&45.0&35.0 &46.6&35.2\\
Gemini-3.0-Flash \cite{gemini3}&\mbluebf{48.9}&45.6&50.8&37.2 &\mbluebf{52.7}&40.9\\
 \noalign{\vspace{-1.5pt}}
 \midrule
 \rowcolor{gray!20}  % 这里添加灰色阴影
 ITIScore (Ours)&\mredbf{51.3}&\mredbf{50.7}&\mredbf{52.6}&\mredbf{42.3}&\mredbf{55.8}&\mredbf{46.5}\\
  \noalign{\vspace{-1.5pt}}
\bottomrule
\end{tabular}
}
\label{multi}
\end{table}

\subsection{Ablation Study}
Table~\ref{ablation} presents an ablation study analyzing the impact of different components and design choices in ITIScore. As shown in Rows 1–2, even without training, ITIScore achieves improvements in the relevance dimension, however, it struggles with multi-dimensional evaluation. Row 3 indicates that fine-tuning the MLLM without incorporating regenerated images leads to suboptimal performance. Rows 4–7 demonstrate that the dimension-wise Gaussian NLL loss consistently outperforms conventional regression losses such as MSE and L1, suggesting that modeling scores as distributions better captures the uncertainty in human annotations. Furthermore, incorporating the aggregate loss further improves performance by enforcing consistency across multiple evaluation dimensions. We also observe that the choice of MLLM backbone has a notable impact on evaluation performance. InternVL3.5 (14B) achieves higher scores on several evaluation dimensions, but these improvements are not consistent across all metrics. In contrast, the 4B variant shows a noticeable drop in performance. InternVL3.5 (8B) provides the best balance, achieving strong correlations with a moderate model size. When comparing models of similar scale, InternVL3.5 consistently outperforms DeepSeekVL2 and Qwen3-VL, demonstrating the advantage of its multimodal representations. Furthermore, replacing the generation model in the image-to-text-to-image framework with FLUX.1 \cite{FLUX} or Stable Diffusion 3 \cite{SD} instead of Qwen-Image \cite{qwenedit} results in slightly lower correlations, highlighting that higher-quality generated images contribute more informative signals for evaluating caption quality. These findings collectively validate the effectiveness of our design choices in ITIScore and the importance of both backbone selection and generation quality.

\subsection{Zero-shot Cross-dataset Evaluation}
We further evaluate the zero-shot cross-dataset generalization of ITIScore on multiple benchmark datasets, including Composites, Flickr8k-CF, Flickr8k-Expert, MMHE, and LongCap-Arena. Tables~\ref{single} and \ref{multi} report the Kendall correlation coefficients between automatic metrics and human judgments across these datasets. ITIScore consistently achieves the highest correlations in both single and multiple dimensional evaluation settings, outperforming traditional reference-based metrics, reference-free metrics, and advanced MLLMs such as ChatGPT-5 and Gemini-3.0-Flash.

\section{Conclusion}
In this paper, we present ICBench, a large-scale dataset for evaluating both short and long captioning capabilities of MLLMs across multiple dimensions. Leveraging ICBench, we propose ITIScore, a novel image-to-text-to-image evaluation metric that assesses caption quality through reconstruction consistency. Extensive experiments show that ITIScore consistently outperforms existing metrics, achieving stronger alignment with human judgments and demonstrating superior cross-dataset generalization.
% TODO

% ===== Acknowledgements (optional) =====
% \begin{acks}
% TODO
% \end{acks}

% ===== References =====
\bibliographystyle{ACM-Reference-Format}
\bibliography{references} % create references.bib (or change to your bib filename)


\begin{thebibliography}{63}

%%% the Web address.

\ifx \showCODEN    \undefined \def \showCODEN     #1{\unskip}     \fi
\ifx \showISBNx    \undefined \def \showISBNx     #1{\unskip}     \fi
\ifx \showISBNxiii \undefined \def \showISBNxiii  #1{\unskip}     \fi
\ifx \showISSN     \undefined \def \showISSN      #1{\unskip}     \fi
\ifx \showLCCN     \undefined \def \showLCCN      #1{\unskip}     \fi
\ifx \shownote     \undefined \def \shownote      #1{#1}          \fi
\ifx \showarticletitle \undefined \def \showarticletitle #1{#1}   \fi
\ifx \showURL      \undefined \def \showURL       {\relax}        \fi
% invisible to TeX
\providecommand\bibfield[2]{#2}
\providecommand\bibinfo[2]{#2}
\providecommand\natexlab[1]{#1}
\providecommand\showeprint[2][]{arXiv:#2}

\bibitem[Aditya et~al\mbox{.}(2015)]%
        {Composite}
\bibfield{author}{\bibinfo{person}{Somak Aditya}, \bibinfo{person}{Yezhou Yang}, \bibinfo{person}{Chitta Baral}, \bibinfo{person}{Cornelia Fermuller}, {and} \bibinfo{person}{Yiannis Aloimonos}.} \bibinfo{year}{2015}\natexlab{}.
\newblock \showarticletitle{From Images to Sentences through Scene Description Graphs using Commonsense Reasoning and Knowledge}.
\newblock \bibinfo{journal}{\emph{arXiv preprint arXiv:1511.03292}} (\bibinfo{year}{2015}).
\newblock


\bibitem[Anderson et~al\mbox{.}(2016)]%
        {spice}
\bibfield{author}{\bibinfo{person}{Peter Anderson}, \bibinfo{person}{Basura Fernando}, \bibinfo{person}{Mark Johnson}, {and} \bibinfo{person}{Stephen Gould}.} \bibinfo{year}{2016}\natexlab{}.
\newblock \showarticletitle{SPICE: Semantic Propositional Image Caption Evaluation}. In \bibinfo{booktitle}{\emph{Proceedings of the European Conference on Computer Vision (ECCV)}}. \bibinfo{pages}{382--398}.
\newblock


\bibitem[Anderson et~al\mbox{.}(2018)]%
        {anderson2018bottom}
\bibfield{author}{\bibinfo{person}{Peter Anderson}, \bibinfo{person}{Xiaodong He}, \bibinfo{person}{Chris Buehler}, \bibinfo{person}{Damien Teney}, \bibinfo{person}{Mark Johnson}, \bibinfo{person}{Stephen Gould}, {and} \bibinfo{person}{Lei Zhang}.} \bibinfo{year}{2018}\natexlab{}.
\newblock \showarticletitle{Bottom-Up and Top-Down Attention for Image Captioning and Visual Question Answering}. In \bibinfo{booktitle}{\emph{Proceedings of the IEEE Conference on Computer Vision and Pattern Recognition (CVPR)}}.
\newblock


\bibitem[Bai et~al\mbox{.}(2025)]%
        {qwenvl2}
\bibfield{author}{\bibinfo{person}{Shuai Bai}, \bibinfo{person}{Keqin Chen}, \bibinfo{person}{Xuejing Liu}, \bibinfo{person}{Jialin Wang}, \bibinfo{person}{Wenbin Ge}, \bibinfo{person}{Sibo Song}, {et~al\mbox{.}}} \bibinfo{year}{2025}\natexlab{}.
\newblock \showarticletitle{Qwen2.5-VL Technical Report}.
\newblock \bibinfo{journal}{\emph{arXiv preprint arXiv:2502.13923}} (\bibinfo{year}{2025}).
\newblock


\bibitem[Banerjee and Lavie(2005)]%
        {banerjee2005meteor}
\bibfield{author}{\bibinfo{person}{Satanjeev Banerjee} {and} \bibinfo{person}{Alon Lavie}.} \bibinfo{year}{2005}\natexlab{}.
\newblock \showarticletitle{METEOR: An Automatic Metric for MT Evaluation with Improved Correlation with Human Judgments}. In \bibinfo{booktitle}{\emph{Proceedings of the ACL Workshop on Intrinsic and Extrinsic Evaluation Measures for Machine Translation and/or Summarization (ACL Workshop)}}.
\newblock


\bibitem[Chan et~al\mbox{.}(2023)]%
        {chan2023clair}
\bibfield{author}{\bibinfo{person}{David Chan}, \bibinfo{person}{Suzanne Petryk}, \bibinfo{person}{Joseph Gonzalez}, \bibinfo{person}{Trevor Darrell}, {and} \bibinfo{person}{John Canny}.} \bibinfo{year}{2023}\natexlab{}.
\newblock \showarticletitle{Clair: Evaluating image captions with large language models}. In \bibinfo{booktitle}{\emph{Proceedings of the 2023 Conference on Empirical Methods in Natural Language Processing (EMNLP)}}. \bibinfo{pages}{13638--13646}.
\newblock


\bibitem[Cheng et~al\mbox{.}(2025)]%
        {cheng2025caparena}
\bibfield{author}{\bibinfo{person}{Kanzhi Cheng}, \bibinfo{person}{Wenpo Song}, \bibinfo{person}{Jiaxin Fan}, \bibinfo{person}{Zheng Ma}, \bibinfo{person}{Qiushi Sun}, \bibinfo{person}{Fangzhi Xu}, {et~al\mbox{.}}} \bibinfo{year}{2025}\natexlab{}.
\newblock \showarticletitle{Caparena: Benchmarking and analyzing detailed image captioning in the llm era}. In \bibinfo{booktitle}{\emph{Proceedings of the Annual Meeting of the Association for Computational Linguistics (ACL)}}. \bibinfo{pages}{14077--14094}.
\newblock


\bibitem[Comanici et~al\mbox{.}(2025)]%
        {nanobanana}
\bibfield{author}{\bibinfo{person}{Gheorghe Comanici}, \bibinfo{person}{Eric Bieber}, \bibinfo{person}{Mike Schaekermann}, \bibinfo{person}{Ice Pasupat}, \bibinfo{person}{Noveen Sachdeva}, \bibinfo{person}{Inderjit Dhillon}, {et~al\mbox{.}}} \bibinfo{year}{2025}\natexlab{}.
\newblock \showarticletitle{Gemini 2.5: Pushing the Frontier with Advanced Reasoning, Multimodality, Long Context, and Next Generation Agentic Capabilities}.
\newblock \bibinfo{journal}{\emph{arXiv preprint arXiv:2507.06261}} (\bibinfo{year}{2025}).
\newblock


\bibitem[Cornia et~al\mbox{.}(2020)]%
        {cornia2020meshed}
\bibfield{author}{\bibinfo{person}{Marcella Cornia}, \bibinfo{person}{Matteo Stefanini}, \bibinfo{person}{Lorenzo Baraldi}, {and} \bibinfo{person}{Rita Cucchiara}.} \bibinfo{year}{2020}\natexlab{}.
\newblock \showarticletitle{Meshed-memory transformer for image captioning}. In \bibinfo{booktitle}{\emph{Proceedings of the IEEE/CVF conference on computer vision and pattern recognition (CVPR)}}. \bibinfo{pages}{10578--10587}.
\newblock


\bibitem[Duan et~al\mbox{.}(2025)]%
        {duan2025finevq}
\bibfield{author}{\bibinfo{person}{Huiyu Duan}, \bibinfo{person}{Qiang Hu}, \bibinfo{person}{Jiarui Wang}, \bibinfo{person}{Liu Yang}, \bibinfo{person}{Zitong Xu}, \bibinfo{person}{Lu Liu}, \bibinfo{person}{Xiongkuo Min}, \bibinfo{person}{Chunlei Cai}, \bibinfo{person}{Tianxiao Ye}, \bibinfo{person}{Xiaoyun Zhang}, {et~al\mbox{.}}} \bibinfo{year}{2025}\natexlab{}.
\newblock \showarticletitle{Finevq: Fine-grained user generated content video quality assessment}. In \bibinfo{booktitle}{\emph{Proceedings of the Computer Vision and Pattern Recognition Conference (CVPR)}}. \bibinfo{pages}{3206--3217}.
\newblock


\bibitem[Duan et~al\mbox{.}(2022)]%
        {duan2022confusing}
\bibfield{author}{\bibinfo{person}{Huiyu Duan}, \bibinfo{person}{Xiongkuo Min}, \bibinfo{person}{Yucheng Zhu}, \bibinfo{person}{Guangtao Zhai}, \bibinfo{person}{Xiaokang Yang}, {and} \bibinfo{person}{Patrick Le~Callet}.} \bibinfo{year}{2022}\natexlab{}.
\newblock \showarticletitle{Confusing image quality assessment: Toward better augmented reality experience}.
\newblock \bibinfo{journal}{\emph{IEEE Transactions on Image Processing (TIP)}}  \bibinfo{volume}{31} (\bibinfo{year}{2022}), \bibinfo{pages}{7206--7221}.
\newblock


\bibitem[Esser et~al\mbox{.}(2024)]%
        {FLUX}
\bibfield{author}{\bibinfo{person}{Patrick Esser}, \bibinfo{person}{Sumith Kulal}, \bibinfo{person}{Andreas Blattmann}, \bibinfo{person}{Rahim Entezari}, \bibinfo{person}{Jonas M\"{u}ller}, \bibinfo{person}{Harry Saini}, {et~al\mbox{.}}} \bibinfo{year}{2024}\natexlab{}.
\newblock \showarticletitle{Scaling rectified flow transformers for high-resolution image synthesis}. In \bibinfo{booktitle}{\emph{Proceedings of the International Conference on Machine Learning (ICML)}}.
\newblock


\bibitem[{Google}(2025)]%
        {gemini3}
\bibfield{author}{\bibinfo{person}{{Google}}.} \bibinfo{year}{2025}\natexlab{}.
\newblock \bibinfo{title}{Gemini 3.0}.
\newblock \bibinfo{howpublished}{\url{https://deepmind.google/technologies/gemini/}}.
\newblock


\bibitem[Hessel et~al\mbox{.}(2021)]%
        {clipscore}
\bibfield{author}{\bibinfo{person}{Jack Hessel}, \bibinfo{person}{Ari Holtzman}, \bibinfo{person}{Maxwell Forbes}, \bibinfo{person}{Ronan Le~Bras}, {and} \bibinfo{person}{Yejin Choi}.} \bibinfo{year}{2021}\natexlab{}.
\newblock \showarticletitle{CLIPScore: A Reference-Free Evaluation Metric for Image Captioning}. In \bibinfo{booktitle}{\emph{Proceedings of the Conference on Empirical Methods in Natural Language Processing (EMNLP)}}. \bibinfo{pages}{7504--7513}.
\newblock


\bibitem[Hodosh et~al\mbox{.}(2013)]%
        {Flickr8K-CF}
\bibfield{author}{\bibinfo{person}{Micah Hodosh}, \bibinfo{person}{Peter Young}, {and} \bibinfo{person}{Julia Hockenmaier}.} \bibinfo{year}{2013}\natexlab{}.
\newblock \showarticletitle{Framing Image Description as a Ranking Task: Data, Models and Evaluation Metrics}.
\newblock \bibinfo{journal}{\emph{Journal of Artificial Intelligence Research (JAIR)}}  \bibinfo{volume}{47} (\bibinfo{year}{2013}), \bibinfo{pages}{853--899}.
\newblock


\bibitem[(ITU)(2012)]%
        {subject}
\bibfield{author}{\bibinfo{person}{International Telecommunication~Union (ITU)}.} \bibinfo{year}{2012}\natexlab{}.
\newblock \bibinfo{booktitle}{\emph{Methodology for the Subjective Assessment of the Quality of Television Pictures}}.
\newblock \bibinfo{type}{{T}echnical {R}eport} Rec. ITU-R BT.500-13. \bibinfo{institution}{International Telecommunication Union (ITU)}.
\newblock


\bibitem[Kasai et~al\mbox{.}(2022)]%
        {kasai2022transparent}
\bibfield{author}{\bibinfo{person}{Jungo Kasai}, \bibinfo{person}{Keisuke Sakaguchi}, \bibinfo{person}{Lavinia Dunagan}, \bibinfo{person}{Jacob Morrison}, \bibinfo{person}{Ronan Le~Bras}, \bibinfo{person}{Yejin Choi}, {and} \bibinfo{person}{Noah~A Smith}.} \bibinfo{year}{2022}\natexlab{}.
\newblock \showarticletitle{Transparent human evaluation for image captioning}. In \bibinfo{booktitle}{\emph{Proceedings of the Conference of the North American Chapter of the Association for Computational Linguistics (NAACL)}}. \bibinfo{pages}{3464--3478}.
\newblock


\bibitem[Kim et~al\mbox{.}(2022)]%
        {kim2022mutual}
\bibfield{author}{\bibinfo{person}{Jin-Hwa Kim}, \bibinfo{person}{Yunji Kim}, \bibinfo{person}{Jiyoung Lee}, \bibinfo{person}{Kang~Min Yoo}, {and} \bibinfo{person}{Sang-Woo Lee}.} \bibinfo{year}{2022}\natexlab{}.
\newblock \showarticletitle{Mutual Information Divergence: A Unified Metric for Multimodal Generative Models}. In \bibinfo{booktitle}{\emph{Advances in Neural Information Processing Systems (NeurIPS)}}. \bibinfo{pages}{35072--35086}.
\newblock


\bibitem[Lee et~al\mbox{.}(2021)]%
        {lee2021umic}
\bibfield{author}{\bibinfo{person}{Hwanhee Lee}, \bibinfo{person}{Seunghyun Yoon}, \bibinfo{person}{Franck Dernoncourt}, \bibinfo{person}{Trung Bui}, {and} \bibinfo{person}{Kyomin Jung}.} \bibinfo{year}{2021}\natexlab{}.
\newblock \showarticletitle{UMIC: An Unreferenced Metric for Image Captioning via Contrastive Learning}. In \bibinfo{booktitle}{\emph{Proceedings of the Annual Meeting of the Association for Computational Linguistics (ACL)}}. \bibinfo{pages}{220--226}.
\newblock


\bibitem[Lee et~al\mbox{.}(2024)]%
        {lee2024fleur}
\bibfield{author}{\bibinfo{person}{Yebin Lee}, \bibinfo{person}{Imseong Park}, {and} \bibinfo{person}{Myungjoo Kang}.} \bibinfo{year}{2024}\natexlab{}.
\newblock \showarticletitle{Fleur: An explainable reference-free evaluation metric for image captioning using a large multimodal model}. In \bibinfo{booktitle}{\emph{Proceedings of the Association for Computational Linguistics (ACL)}}. \bibinfo{pages}{3732--3746}.
\newblock


\bibitem[Li et~al\mbox{.}(2022)]%
        {li2022blip}
\bibfield{author}{\bibinfo{person}{Junnan Li}, \bibinfo{person}{Dongxu Li}, \bibinfo{person}{Caiming Xiong}, {and} \bibinfo{person}{Steven C.~H. Hoi}.} \bibinfo{year}{2022}\natexlab{}.
\newblock \showarticletitle{BLIP: Bootstrapping Language-Image Pre-training for Unified Vision-Language Understanding and Generation}. In \bibinfo{booktitle}{\emph{Proceedings of the International Conference on Machine Learning (ICML)}}.
\newblock


\bibitem[Lin(2004)]%
        {lin2004rouge}
\bibfield{author}{\bibinfo{person}{Chin-Yew Lin}.} \bibinfo{year}{2004}\natexlab{}.
\newblock \showarticletitle{ROUGE: A Package for Automatic Evaluation of Summaries}. In \bibinfo{booktitle}{\emph{Text Summarization Branches Out: Proceedings of the ACL Workshop}}.
\newblock


\bibitem[Liu et~al\mbox{.}(2024a)]%
        {deepseekv2}
\bibfield{author}{\bibinfo{person}{Aixin Liu}, \bibinfo{person}{Bei Feng}, \bibinfo{person}{Bin Wang}, \bibinfo{person}{Bingxuan Wang}, \bibinfo{person}{Bo Liu}, \bibinfo{person}{Chenggang Zhao}, {et~al\mbox{.}}} \bibinfo{year}{2024}\natexlab{a}.
\newblock \showarticletitle{DeepSeek-V2: A Strong, Economical, and Efficient Mixture-of-Experts Language Model}.
\newblock \bibinfo{journal}{\emph{arXiv preprint arXiv:2405.04434}} (\bibinfo{year}{2024}).
\newblock


\bibitem[Liu et~al\mbox{.}(2024b)]%
        {llava}
\bibfield{author}{\bibinfo{person}{Haotian Liu}, \bibinfo{person}{Chunyuan Li}, \bibinfo{person}{Yuheng Li}, {and} \bibinfo{person}{Yong~Jae Lee}.} \bibinfo{year}{2024}\natexlab{b}.
\newblock \showarticletitle{Improved Baselines with Visual Instruction Tuning}. In \bibinfo{booktitle}{\emph{Proceedings of the IEEE/CVF Conference on Computer Vision and Pattern Recognition (CVPR)}}. \bibinfo{pages}{26296--26306}.
\newblock


\bibitem[Liu et~al\mbox{.}(2025)]%
        {liu2025moa}
\bibfield{author}{\bibinfo{person}{Lu Liu}, \bibinfo{person}{Chunlei Cai}, \bibinfo{person}{Shaocheng Shen}, \bibinfo{person}{Jianfeng Liang}, \bibinfo{person}{Weimin Ouyang}, \bibinfo{person}{Tianxiao Ye}, \bibinfo{person}{Jian Mao}, \bibinfo{person}{Huiyu Duan}, \bibinfo{person}{Jiangchao Yao}, \bibinfo{person}{Xiaoyun Zhang}, {et~al\mbox{.}}} \bibinfo{year}{2025}\natexlab{}.
\newblock \showarticletitle{MoA-VR: A Mixture-of-Agents System Towards All-in-One Video Restoration}.
\newblock \bibinfo{journal}{\emph{IEEE Journal of Selected Topics in Signal Processing}} (\bibinfo{year}{2025}).
\newblock


\bibitem[Liu et~al\mbox{.}(2024c)]%
        {vela}
\bibfield{author}{\bibinfo{person}{Xin Liu}, \bibinfo{person}{Yang Li}, \bibinfo{person}{Zhe Wang}, {et~al\mbox{.}}} \bibinfo{year}{2024}\natexlab{c}.
\newblock \showarticletitle{VELA: Evaluating Long-form Image Captioning with Automated Metrics and Human Feedback}. In \bibinfo{booktitle}{\emph{Proceedings of the IEEE/CVF Conference on Computer Vision and Pattern Recognition (CVPR)}}.
\newblock


\bibitem[Lu et~al\mbox{.}(2025)]%
        {ovis25}
\bibfield{author}{\bibinfo{person}{Shiyin Lu}, \bibinfo{person}{Yang Li}, \bibinfo{person}{Yu Xia}, \bibinfo{person}{Yuwei Hu}, \bibinfo{person}{Shanshan Zhao}, \bibinfo{person}{Yanqing Ma}, {et~al\mbox{.}}} \bibinfo{year}{2025}\natexlab{}.
\newblock \showarticletitle{Ovis2.5 Technical Report}.
\newblock \bibinfo{journal}{\emph{arXiv:2508.11737}} (\bibinfo{year}{2025}).
\newblock


\bibitem[Maeda et~al\mbox{.}(2024)]%
        {maeda2024vision}
\bibfield{author}{\bibinfo{person}{Koki Maeda}, \bibinfo{person}{Shuhei Kurita}, \bibinfo{person}{Taiki Miyanishi}, {and} \bibinfo{person}{Naoaki Okazaki}.} \bibinfo{year}{2024}\natexlab{}.
\newblock \showarticletitle{Vision language model-based caption evaluation method leveraging visual context extraction}.
\newblock \bibinfo{journal}{\emph{arXiv preprint arXiv:2402.17969}} (\bibinfo{year}{2024}).
\newblock


\bibitem[Meta(2024)]%
        {llama}
\bibfield{author}{\bibinfo{person}{AI Meta}.} \bibinfo{year}{2024}\natexlab{}.
\newblock \showarticletitle{Llama 3.2: Revolutionizing Edge AI and Vision with Open, Customizable Models}.
\newblock \bibinfo{journal}{\emph{Meta AI Blog. Retrieved December}} (\bibinfo{year}{2024}).
\newblock


\bibitem[Ohi et~al\mbox{.}(2024)]%
        {ohi2024multi}
\bibfield{author}{\bibinfo{person}{Masanari Ohi}, \bibinfo{person}{Masahiro Kaneko}, \bibinfo{person}{Naoaki Okazaki}, {and} \bibinfo{person}{Nakamasa Inoue}.} \bibinfo{year}{2024}\natexlab{}.
\newblock \showarticletitle{Multi-modal, Multi-task, Multi-criteria Automatic Evaluation with Vision Language Models}.
\newblock \bibinfo{journal}{\emph{arXiv preprint arXiv:2412.14613}} (\bibinfo{year}{2024}).
\newblock


\bibitem[OpenAI(2025)]%
        {chatgpt4o}
\bibfield{author}{\bibinfo{person}{OpenAI}.} \bibinfo{year}{2025}\natexlab{}.
\newblock \bibinfo{title}{ChatGPT-4o: Advanced Multimodal Chat Model}.
\newblock
\urldef\tempurl%
\url{https://openai.com/chatgpt}
\showURL{%
\tempurl}


\bibitem[{OpenAI}(2025)]%
        {gpt5}
\bibfield{author}{\bibinfo{person}{{OpenAI}}.} \bibinfo{year}{2025}\natexlab{}.
\newblock \bibinfo{title}{GPT-5}.
\newblock \bibinfo{howpublished}{\url{https://www.openai.com}}.
\newblock


\bibitem[Pan et~al\mbox{.}(2020)]%
        {pan2020x}
\bibfield{author}{\bibinfo{person}{Yingwei Pan}, \bibinfo{person}{Ting Yao}, \bibinfo{person}{Yehao Li}, {and} \bibinfo{person}{Tao Mei}.} \bibinfo{year}{2020}\natexlab{}.
\newblock \showarticletitle{X-linear attention networks for image captioning}. In \bibinfo{booktitle}{\emph{Proceedings of the IEEE/CVF conference on computer vision and pattern recognition (CVPR)}}. \bibinfo{pages}{10971--10980}.
\newblock


\bibitem[Papineni et~al\mbox{.}(2002)]%
        {papineni2002bleu}
\bibfield{author}{\bibinfo{person}{Kishore Papineni}, \bibinfo{person}{Salim Roukos}, \bibinfo{person}{Todd Ward}, {and} \bibinfo{person}{Wei-Jing Zhu}.} \bibinfo{year}{2002}\natexlab{}.
\newblock \showarticletitle{BLEU: a Method for Automatic Evaluation of Machine Translation}. In \bibinfo{booktitle}{\emph{Proceedings of the Annual Meeting of the Association for Computational Linguistics (ACL)}}.
\newblock


\bibitem[Radford et~al\mbox{.}(2021)]%
        {radford2021clip}
\bibfield{author}{\bibinfo{person}{Alec Radford}, \bibinfo{person}{Jong~Wook Kim}, \bibinfo{person}{Chris Hallacy}, \bibinfo{person}{Aditya Ramesh}, \bibinfo{person}{Gabriel Goh}, \bibinfo{person}{Sandhini Agarwal}, \bibinfo{person}{Girish Sastry}, \bibinfo{person}{Amanda Askell}, \bibinfo{person}{Pamela Mishkin}, \bibinfo{person}{Jack Clark}, \bibinfo{person}{Gretchen Krueger}, {and} \bibinfo{person}{Ilya Sutskever}.} \bibinfo{year}{2021}\natexlab{}.
\newblock \showarticletitle{Learning Transferable Visual Models From Natural Language Supervision}. In \bibinfo{booktitle}{\emph{Proceedings of the International Conference on Machine Learning (ICML)}}.
\newblock


\bibitem[Rombach et~al\mbox{.}(2022)]%
        {SD}
\bibfield{author}{\bibinfo{person}{Robin Rombach}, \bibinfo{person}{Andreas Blattmann}, \bibinfo{person}{Dominik Lorenz}, \bibinfo{person}{Patrick Esser}, {and} \bibinfo{person}{Bj{\"o}rn Ommer}.} \bibinfo{year}{2022}\natexlab{}.
\newblock \showarticletitle{High-resolution image synthesis with latent diffusion models}. In \bibinfo{booktitle}{\emph{Proceedings of the IEEE/CVF Conference on Computer Vision and Pattern Recognition (CVPR)}}. \bibinfo{pages}{10684--10695}.
\newblock


\bibitem[Rotstein et~al\mbox{.}(2024)]%
        {rotstein2024fusecap}
\bibfield{author}{\bibinfo{person}{Noam Rotstein}, \bibinfo{person}{David Bensaid}, \bibinfo{person}{Shaked Brody}, \bibinfo{person}{Roy Ganz}, {and} \bibinfo{person}{Ron Kimmel}.} \bibinfo{year}{2024}\natexlab{}.
\newblock \showarticletitle{Fusecap: Leveraging large language models for enriched fused image captions}. In \bibinfo{booktitle}{\emph{Proceedings of the IEEE/CVF winter conference on applications of computer vision (CVPR)}}. \bibinfo{pages}{5689--5700}.
\newblock


\bibitem[Sarto et~al\mbox{.}(2023)]%
        {sarto2023positive}
\bibfield{author}{\bibinfo{person}{Sara Sarto}, \bibinfo{person}{Manuele Barraco}, \bibinfo{person}{Marcella Cornia}, \bibinfo{person}{Lorenzo Baraldi}, {and} \bibinfo{person}{Rita Cucchiara}.} \bibinfo{year}{2023}\natexlab{}.
\newblock \showarticletitle{Positive-Augmented Contrastive Learning for Image and Video Captioning Evaluation}. In \bibinfo{booktitle}{\emph{Proceedings of the IEEE/CVF Conference on Computer Vision and Pattern Recognition (CVPR)}}. \bibinfo{pages}{6914--6924}.
\newblock


\bibitem[Sarto et~al\mbox{.}(2024)]%
        {sarto2024bridge}
\bibfield{author}{\bibinfo{person}{Sara Sarto}, \bibinfo{person}{Marcella Cornia}, \bibinfo{person}{Lorenzo Baraldi}, {and} \bibinfo{person}{Rita Cucchiara}.} \bibinfo{year}{2024}\natexlab{}.
\newblock \showarticletitle{BRIDGE: Bridging gaps in image captioning evaluation with stronger visual cues}. In \bibinfo{booktitle}{\emph{Proceedings of the European Conference on Computer Vision (ECCV)}}. \bibinfo{pages}{70--87}.
\newblock


\bibitem[Sarto et~al\mbox{.}(2025)]%
        {sarto2025image}
\bibfield{author}{\bibinfo{person}{Sara Sarto}, \bibinfo{person}{Marcella Cornia}, {and} \bibinfo{person}{Rita Cucchiara}.} \bibinfo{year}{2025}\natexlab{}.
\newblock \showarticletitle{Image captioning evaluation in the age of multimodal llms: Challenges and future perspectives}.
\newblock \bibinfo{journal}{\emph{arXiv preprint arXiv:2503.14604}} (\bibinfo{year}{2025}).
\newblock


\bibitem[Tong et~al\mbox{.}(2025)]%
        {tong2025g}
\bibfield{author}{\bibinfo{person}{Tony~Cheng Tong}, \bibinfo{person}{Sirui He}, \bibinfo{person}{Zhiwen Shao}, {and} \bibinfo{person}{Dit-Yan Yeung}.} \bibinfo{year}{2025}\natexlab{}.
\newblock \showarticletitle{G-veval: A versatile metric for evaluating image and video captions using gpt-4o}. In \bibinfo{booktitle}{\emph{Proceedings of the Conference on Association for the Advancement of Artificial Intelligence (AAAI)}}, Vol.~\bibinfo{volume}{39}. \bibinfo{pages}{7419--7427}.
\newblock


\bibitem[Vedantam et~al\mbox{.}(2015)]%
        {cider}
\bibfield{author}{\bibinfo{person}{Ramakrishna Vedantam}, \bibinfo{person}{C.~Lawrence Zitnick}, {and} \bibinfo{person}{Devi Parikh}.} \bibinfo{year}{2015}\natexlab{}.
\newblock \showarticletitle{CIDEr: Consensus-based Image Description Evaluation}. In \bibinfo{booktitle}{\emph{Proceedings of the IEEE Conference on Computer Vision and Pattern Recognition (CVPR)}}. \bibinfo{pages}{4566--4575}.
\newblock


\bibitem[Vinyals et~al\mbox{.}(2015)]%
        {vinyals2015show}
\bibfield{author}{\bibinfo{person}{Oriol Vinyals}, \bibinfo{person}{Alexander Toshev}, \bibinfo{person}{Samy Bengio}, {and} \bibinfo{person}{Dumitru Erhan}.} \bibinfo{year}{2015}\natexlab{}.
\newblock \showarticletitle{Show and Tell: A Neural Image Caption Generator}. In \bibinfo{booktitle}{\emph{Proceedings of the IEEE Conference on Computer Vision and Pattern Recognition (CVPR)}}.
\newblock


\bibitem[Wada et~al\mbox{.}(2024)]%
        {polaris}
\bibfield{author}{\bibinfo{person}{Yuiga Wada}, \bibinfo{person}{Kanta Kaneda}, \bibinfo{person}{Daichi Saito}, {and} \bibinfo{person}{Komei Sugiura}.} \bibinfo{year}{2024}\natexlab{}.
\newblock \showarticletitle{Polos: Multimodal metric learning from human feedback for image captioning}. In \bibinfo{booktitle}{\emph{Proceedings of the IEEE/CVF Conference on Computer Vision and Pattern Recognition (CVPR)}}. \bibinfo{pages}{13559--13568}.
\newblock


\bibitem[Wang et~al\mbox{.}(2020)]%
        {wang2020overview}
\bibfield{author}{\bibinfo{person}{Haoran Wang}, \bibinfo{person}{Yue Zhang}, {and} \bibinfo{person}{Xiaosheng Yu}.} \bibinfo{year}{2020}\natexlab{}.
\newblock \showarticletitle{An overview of image caption generation methods}.
\newblock \bibinfo{journal}{\emph{Computational intelligence and neuroscience (CIN)}} \bibinfo{volume}{2020}, \bibinfo{number}{1} (\bibinfo{year}{2020}), \bibinfo{pages}{3062706}.
\newblock


\bibitem[Wang et~al\mbox{.}(2026)]%
        {wang2026quality}
\bibfield{author}{\bibinfo{person}{Jiarui Wang}, \bibinfo{person}{Huiyu Duan}, \bibinfo{person}{Guangtao Zhai}, {and} \bibinfo{person}{Xiongkuo Min}.} \bibinfo{year}{2026}\natexlab{}.
\newblock \showarticletitle{Quality assessment for AI generated images with instruction tuning}.
\newblock \bibinfo{journal}{\emph{IEEE Transactions on Multimedia (TMM)}} (\bibinfo{year}{2026}).
\newblock


\bibitem[Wang et~al\mbox{.}(2025a)]%
        {wang2025lmm4lmm}
\bibfield{author}{\bibinfo{person}{Jiarui Wang}, \bibinfo{person}{Huiyu Duan}, \bibinfo{person}{Yu Zhao}, \bibinfo{person}{Juntong Wang}, \bibinfo{person}{Guangtao Zhai}, {and} \bibinfo{person}{Xiongkuo Min}.} \bibinfo{year}{2025}\natexlab{a}.
\newblock \showarticletitle{Lmm4lmm: Benchmarking and evaluating large-multimodal image generation with lmms}. In \bibinfo{booktitle}{\emph{Proceedings of the IEEE/CVF International Conference on Computer Vision (ICCV)}}. \bibinfo{pages}{17312--17323}.
\newblock


\bibitem[Wang et~al\mbox{.}(2022)]%
        {wang2022git}
\bibfield{author}{\bibinfo{person}{Jianfeng Wang}, \bibinfo{person}{Zhengyuan Yang}, \bibinfo{person}{Xiaowei Hu}, \bibinfo{person}{Linjie Li}, \bibinfo{person}{Kevin Lin}, \bibinfo{person}{Zhe Gan}, \bibinfo{person}{Zicheng Liu}, \bibinfo{person}{Ce Liu}, {and} \bibinfo{person}{Lijuan Wang}.} \bibinfo{year}{2022}\natexlab{}.
\newblock \showarticletitle{Git: A generative image-to-text transformer for vision and language}.
\newblock \bibinfo{journal}{\emph{arXiv preprint arXiv:2205.14100}} (\bibinfo{year}{2022}).
\newblock


\bibitem[Wang et~al\mbox{.}(2025b)]%
        {internvl3_5}
\bibfield{author}{\bibinfo{person}{Weiyun Wang}, \bibinfo{person}{Zhangwei Gao}, \bibinfo{person}{Lixin Gu}, \bibinfo{person}{Hengjun Pu}, \bibinfo{person}{Long Cui}, \bibinfo{person}{Xingguang Wei}, {et~al\mbox{.}}} \bibinfo{year}{2025}\natexlab{b}.
\newblock \showarticletitle{InternVL3.5: Advancing Open-Source Multimodal Models in Versatility, Reasoning, and Efficiency}.
\newblock \bibinfo{journal}{\emph{arXiv preprint arXiv:2508.18265}} (\bibinfo{year}{2025}).
\newblock


\bibitem[Wu et~al\mbox{.}(2025)]%
        {qwenedit}
\bibfield{author}{\bibinfo{person}{Chenfei Wu}, \bibinfo{person}{Jiahao Li}, \bibinfo{person}{Jingren Zhou}, \bibinfo{person}{Junyang Lin}, \bibinfo{person}{Kaiyuan Gao}, \bibinfo{person}{Kun Yan}, {et~al\mbox{.}}} \bibinfo{year}{2025}\natexlab{}.
\newblock \showarticletitle{Qwen-Image Technical Report}.
\newblock \bibinfo{journal}{\emph{arXiv preprint arXiv:2508.02324}} (\bibinfo{year}{2025}).
\newblock


\bibitem[Xu et~al\mbox{.}(2015)]%
        {xu2015show}
\bibfield{author}{\bibinfo{person}{Kelvin Xu}, \bibinfo{person}{Jimmy Ba}, \bibinfo{person}{Ryan Kiros}, \bibinfo{person}{Kyunghyun Cho}, \bibinfo{person}{Aaron Courville}, \bibinfo{person}{Ruslan Salakhutdinov}, \bibinfo{person}{Richard Zemel}, {and} \bibinfo{person}{Yoshua Bengio}.} \bibinfo{year}{2015}\natexlab{}.
\newblock \showarticletitle{Show, Attend and Tell: Neural Image Caption Generation with Visual Attention}. In \bibinfo{booktitle}{\emph{Proceedings of the International Conference on Machine Learning (ICML)}}.
\newblock


\bibitem[Xu et~al\mbox{.}(2023)]%
        {xu2023deep}
\bibfield{author}{\bibinfo{person}{Liming Xu}, \bibinfo{person}{Quan Tang}, \bibinfo{person}{Jiancheng Lv}, \bibinfo{person}{Bochuan Zheng}, \bibinfo{person}{Xianhua Zeng}, {and} \bibinfo{person}{Weisheng Li}.} \bibinfo{year}{2023}\natexlab{}.
\newblock \showarticletitle{Deep image captioning: A review of methods, trends and future challenges}.
\newblock \bibinfo{journal}{\emph{Neurocomputing}}  \bibinfo{volume}{546} (\bibinfo{year}{2023}), \bibinfo{pages}{126287}.
\newblock


\bibitem[Xu et~al\mbox{.}(2026a)]%
        {xu2026edithf1mmillionscalerichhuman}
\bibfield{author}{\bibinfo{person}{Zitong Xu}, \bibinfo{person}{Huiyu Duan}, \bibinfo{person}{Zhongpeng Ji}, \bibinfo{person}{Xinyun Zhang}, \bibinfo{person}{Yutao Liu}, \bibinfo{person}{Xiongkuo Min}, {et~al\mbox{.}}} \bibinfo{year}{2026}\natexlab{a}.
\newblock \showarticletitle{EditHF-1M: A Million-Scale Rich Human Preference Feedback for Image Editing}.
\newblock \bibinfo{journal}{\emph{arXiv preprint arXiv:2603.14916}} (\bibinfo{year}{2026}).
\newblock


\bibitem[Xu et~al\mbox{.}(2025a)]%
        {lmm4edit}
\bibfield{author}{\bibinfo{person}{Zitong Xu}, \bibinfo{person}{Huiyu Duan}, \bibinfo{person}{Bingnan Liu}, \bibinfo{person}{Guangji Ma}, \bibinfo{person}{Jiarui Wang}, \bibinfo{person}{Liu Yang}, {et~al\mbox{.}}} \bibinfo{year}{2025}\natexlab{a}.
\newblock \showarticletitle{LMM4Edit: Benchmarking and Evaluating Multimodal Image Editing with LMMs}. In \bibinfo{booktitle}{\emph{Proceedings of the ACM International Conference on Multimedia (ACM MM)}}. \bibinfo{pages}{6908--6917}.
\newblock


\bibitem[Xu et~al\mbox{.}(2025b)]%
        {harmonyiqa}
\bibfield{author}{\bibinfo{person}{Zitong Xu}, \bibinfo{person}{Huiyu Duan}, \bibinfo{person}{Guangji Ma}, \bibinfo{person}{Liu Yang}, \bibinfo{person}{Jiarui Wang}, \bibinfo{person}{Qingbo Wu}, {et~al\mbox{.}}} \bibinfo{year}{2025}\natexlab{b}.
\newblock \showarticletitle{Harmonyiqa: Pioneering benchmark and model for image harmonization quality assessment}. In \bibinfo{booktitle}{\emph{IEEE International Conference on Multimedia and Expo (ICME)}}. \bibinfo{pages}{1--6}.
\newblock


\bibitem[Xu et~al\mbox{.}(2026b)]%
        {xu2026manipshieldunifiedframeworkimage}
\bibfield{author}{\bibinfo{person}{Zitong Xu}, \bibinfo{person}{Huiyu Duan}, \bibinfo{person}{Xiaoyu Wang}, \bibinfo{person}{Zhaolin Cai}, \bibinfo{person}{Kaiwei Zhang}, \bibinfo{person}{Qiang Hu}, \bibinfo{person}{Jing Liu}, \bibinfo{person}{Xiongkuo Min}, {and} \bibinfo{person}{Guangtao Zhai}.} \bibinfo{year}{2026}\natexlab{b}.
\newblock \showarticletitle{ManipShield: A Unified Framework for Image Manipulation Detection, Localization and Explanation}.
\newblock \bibinfo{journal}{\emph{arXiv preprint arXiv:2511.14259}} (\bibinfo{year}{2026}).
\newblock


\bibitem[Yang et~al\mbox{.}(2025b)]%
        {qwen3}
\bibfield{author}{\bibinfo{person}{An Yang}, \bibinfo{person}{Anfeng Li}, \bibinfo{person}{Baosong Yang}, \bibinfo{person}{Beichen Zhang}, \bibinfo{person}{Binyuan Hui}, \bibinfo{person}{Bo Zheng}, {et~al\mbox{.}}} \bibinfo{year}{2025}\natexlab{b}.
\newblock \showarticletitle{Qwen3 Technical Report}.
\newblock \bibinfo{journal}{\emph{arXiv preprint arXiv:2505.09388}} (\bibinfo{year}{2025}).
\newblock


\bibitem[Yang et~al\mbox{.}(2025a)]%
        {yang2025odi}
\bibfield{author}{\bibinfo{person}{Liu Yang}, \bibinfo{person}{Huiyu Duan}, \bibinfo{person}{Ran Tao}, \bibinfo{person}{Juntao Cheng}, \bibinfo{person}{Sijing Wu}, \bibinfo{person}{Yunhao Li}, {et~al\mbox{.}}} \bibinfo{year}{2025}\natexlab{a}.
\newblock \showarticletitle{ODI-Bench: Can MLLMs Understand Immersive Omnidirectional Environments?}
\newblock \bibinfo{journal}{\emph{arXiv preprint arXiv:2510.11549}} (\bibinfo{year}{2025}).
\newblock


\bibitem[Yao et~al\mbox{.}(2024b)]%
        {minicpm}
\bibfield{author}{\bibinfo{person}{Yuan Yao}, \bibinfo{person}{Tianyu Yu}, \bibinfo{person}{Ao Zhang}, \bibinfo{person}{Chongyi Wang}, \bibinfo{person}{Junbo Cui}, \bibinfo{person}{Hongji Zhu}, {et~al\mbox{.}}} \bibinfo{year}{2024}\natexlab{b}.
\newblock \showarticletitle{MiniCPM-V: A GPT-4V Level MLLM on Your Phone}.
\newblock \bibinfo{journal}{\emph{arXiv preprint arXiv:2408.01800}} (\bibinfo{year}{2024}).
\newblock


\bibitem[Yao et~al\mbox{.}(2024a)]%
        {yao2024hifi}
\bibfield{author}{\bibinfo{person}{Ziwei Yao}, \bibinfo{person}{Ruiping Wang}, {and} \bibinfo{person}{Xilin Chen}.} \bibinfo{year}{2024}\natexlab{a}.
\newblock \showarticletitle{Hifi-score: Fine-grained image description evaluation with hierarchical parsing graphs}. In \bibinfo{booktitle}{\emph{Proceedings of the European Conference on Computer Vision (ECCV)}}. \bibinfo{pages}{441--458}.
\newblock


\bibitem[Ye et~al\mbox{.}(2024)]%
        {mplug}
\bibfield{author}{\bibinfo{person}{Jiabo Ye}, \bibinfo{person}{Haiyang Xu}, \bibinfo{person}{Haowei Liu}, \bibinfo{person}{Anwen Hu}, \bibinfo{person}{Ming Yan}, \bibinfo{person}{Qi Qian}, {et~al\mbox{.}}} \bibinfo{year}{2024}\natexlab{}.
\newblock \showarticletitle{mPLUG-Owl3: Towards Long Image-Sequence Understanding in Multimodal Large Language Models}. In \bibinfo{booktitle}{\emph{Proceedings of the International Conference on Learning Representations (ICLR)}}.
\newblock


\bibitem[Zhang et~al\mbox{.}(2020)]%
        {zhang2020bertscore}
\bibfield{author}{\bibinfo{person}{Tianyi Zhang}, \bibinfo{person}{Varsha Kishore}, \bibinfo{person}{Felix Wu}, \bibinfo{person}{Kilian~Q. Weinberger}, {and} \bibinfo{person}{Yoav Artzi}.} \bibinfo{year}{2020}\natexlab{}.
\newblock \showarticletitle{BERTScore: Evaluating Text Generation with BERT}.
\newblock \bibinfo{journal}{\emph{arXiv preprint arXiv:1904.09675}} (\bibinfo{year}{2020}).
\newblock


\bibitem[Zhu et~al\mbox{.}(2025)]%
        {internvl3}
\bibfield{author}{\bibinfo{person}{Jinguo Zhu}, \bibinfo{person}{Weiyun Wang}, \bibinfo{person}{Zhe Chen}, \bibinfo{person}{Zhaoyang Liu}, \bibinfo{person}{Shenglong Ye}, \bibinfo{person}{Lixin Gu}, {et~al\mbox{.}}} \bibinfo{year}{2025}\natexlab{}.
\newblock \showarticletitle{InternVL3: Exploring Advanced Training and Test-Time Recipes for Open-Source Multimodal Models}.
\newblock \bibinfo{journal}{\emph{arXiv preprint arXiv:2504.10479}} (\bibinfo{year}{2025}).
\newblock


\end{thebibliography}

% ===== Appendix (optional) =====
% \appendix
% \section{Additional Details}
% TODO

\end{document}